%% file: collas2025_conference.tex
\documentclass{article} 
\usepackage{collas2025_conference,times}
\usepackage{easyReview}

\input{math_commands.tex}

\usepackage{booktabs} 
\usepackage{amsmath}  
\usepackage{graphicx}
\usepackage{hyperref}   
\usepackage{subcaption}
\usepackage{xcolor}
\usepackage{algorithm}
\usepackage{algpseudocode}
\hypersetup{
    colorlinks=true,
    linkcolor=red,
    filecolor=magenta,
    urlcolor=blue,
    citecolor=purple,
    pdftitle={Overleaf Example},
    pdfpagemode=FullScreen,
    }

\title{Revisiting Replay and Gradient Alignment for \\ Continual Pre-Training of Large Language Models}
\vspace{8em}
\author{
\parbox{\textwidth}{
\centering
\vspace{2em}
Istabrak Abbes$^{1,2,3}$\thanks{Equal contribution (Alphabetical Order)}\quad
Gopeshh Subbaraj$^{1,2}$\footnotemark[1]\quad
Matthew Riemer$^{1,2,4}$\quad
Nizar Islah$^{1,2}$\quad
Benjamin Thérien$^{1,2}$\quad
Tsuguchika Tabaru$^5$\quad
Hiroaki Kingetsu$^5$\quad
Sarath Chandar$^{2,3,6}$\quad
Irina Rish$^{1,2}$\\[1em]
\normalfont
$^1$Universit\'e de Montr\'eal \qquad
$^2$Mila – Quebec AI Institute \qquad
$^3$Chandar Research Lab\\
$^4$IBM Research \qquad
$^5$Fujitsu Research \qquad
$^6$Polytechnique Montr\'eal\\[0.75em]
\texttt{\{istabrak.abbes, gopeshh.subbaraj\}@mila.quebec}
}
}

\collasfinalcopy 
\begin{document}
\maketitle
\begin{abstract}
Training large language models (LLMs) typically involves pre-training on massive corpora, only to restart the process entirely when new data becomes available. 
A more efficient and resource-conserving approach would be continual pre-training, where models are updated with new data rather than retraining from scratch. However, the introduction of new data often causes distribution shifts, leading to performance degradation on previously learned tasks. 
In this paper, we take a deeper look at two popular proposals for addressing this distribution shift within the continual learning literature: experience replay and gradient alignment. We consider continual pre-training of models within the Llama family of architectures at a large scale across languages with 100 billion tokens of training data in each language, finding that both replay and gradient alignment lead to more stable learning without forgetting. This conclusion holds both as we vary the model scale and as we vary the number and diversity of tasks. Moreover, we are the first to demonstrate the effectiveness of gradient alignment techniques in the context of LLM pre-training and propose an efficient implementation of meta-experience replay (MER) \citep{riemer2019learninglearnforgettingmaximizing} that imbues experience replay with the benefits of gradient alignment despite negligible compute and memory overhead. Our scaling analysis across model sizes and replay rates indicates that small rates of replaying old examples are definitely a more valuable use of compute than investing in model size, but that it is more compute efficient to scale the size of the model than invest in high rates of replaying old examples.

\end{abstract}
\vspace{-0.4cm}

\section{Introduction}
\vspace{-0.2cm}
Large Language Models (LLMs) need regular updates to be current with new information and domains, posing a problem for organizations looking to maintain LLMs without repeatedly performing expensive retraining from scratch. Performing updates to a model that has already received pre-training on a new distribution is the classic problem of \textit{continual learning} \citep{Ring94} or \textit{lifelong learning} \citep{Thrun94}. Unfortunately, achieving good performance in these non-i.i.d. settings is famously difficult because of the \textit{stability-plasticity dilemma} \citep{StabilityPlasticity}, formalizing the inherent tension between achieving performance improvements on new data and maintaining performance on old data. In this paper, we are concerned with the setting termed \textit{continual pre-training} \citep{ke2023continual,gupta2023continualpretraininglargelanguage,ibrahim2024simple,roth2024practitioner,chen2024effectiveefficientcontinualpretraining,fujii2024continual,li2024examining,zhou2024continual} in which a pre-trainined LLM is continually trained on a large amount of new data (i.e. 100 billion tokens). We should draw a strong distinction between this setting and other settings such as fine-tuning or instruction tuning, which are generally characterized by training on much smaller datasets for a much smaller number of gradient steps. This is an important difference because the risk of experiencing the problem of \textit{catastrophic forgetting} \citep{CF}, in which much of a model's old knowledge is overwritten as it is trying to learn new knowledge, grows as the number of training steps on the new distribution are increased. This makes effective continual pre-training very difficult to achieve and less common in practice -- despite the profound opportunity it would present for many organizations in managing their compute costs and carbon footprint if it were successful. 


One of the most generally applicable and successful approaches for stabilizing continual learning is \textit{experience replay} \citep{Lin92,Murre92,Robins95}. The main idea is to store a buffer of past experiences that are interleaved with incoming experiences to stabilize learning. While the distribution of incoming data may be non-nonstationary, the network can optimize for the stationary distribution over examples seen so far with replay. As such, replay provides a general solution to the stability-plasticity dilemma by allowing neural networks to retain more past information in exchange for compute. While the benefits of replay are well established in the literature, it is less clear whether replay is always the most efficient way of investing in compute. In our experiments, we directly compare replay with models that invest the same amount of computation in more parameters, finding evidence that replay can be a wiser investment. In our paper, we are mostly concerned with the amount of compute in floating point operations (FLOPs) or VRAM in bytes that a given model utilizes as we believe these are the main constraints in scaling LLM pre-training. Much of the continual learning literature focuses on the size of the buffer used for replay \citep{gem,Recollections,riemer2019learninglearnforgettingmaximizing,agem,chaudhry2019tinyepisodicmemoriescontinual}. However, it is possible to efficiently implement storage of this buffer on disk, so we do not consider any restrictions on the replay buffer size in our experiments over 500 billion tokens. 


That said, if we want to utilize compute in the most efficient way possible to address the stability-plasticity dilemma, it is tempting to consider improvements to replay that can enhance the efficiency of learning. One prominent research direction looking to achieve just that is papers on promoting \textit{gradient alignment} during continual learning. The idea is that at any instant in time during gradient based learning, training on one data point will experience transfer to the other data point if the dot product between the gradients with respect to both points is positive and experience interference to or forgetting of the other data point if the dot product between the gradients with respect to both points is negative. As such, approaches including Gradient Episodic Memory (GEM) \citep{gem}, its efficient approximation \citep{agem}, and PCGrad \citep{pcgrad} consider a constrained optimization procedure in which learning is allowed on new data only if it doesn't interfere with past data. This idea was generalized by \citet{riemer2019learninglearnforgettingmaximizing} as a regularized learning objective driven by optimization based meta-learning \citep{maml,nichol2018firstordermetalearningalgorithms} whereby learning is regularized to also maximize the dot product between gradients such that not only is interference minimized, but transfer is also maximized. Another advantage of the meta-learning perspective is that the network can learn to shift its distribution of gradient dot products over the course of learning rather than performing adhoc gradient projections only associated with a single batch \citep{riemer2019learninglearnforgettingmaximizing}. In our paper, we are the first to adapt the \textit{meta-experience replay} approach of \citet{riemer2019learninglearnforgettingmaximizing} for efficient use the context of modern LLMs. This is the first showcase of gradient alignment approaches in the context of continual pre-training with LLMs. 






In this paper, we focus on a realistic scenario of sequential domain adaptation: a model is first pretrained on an English corpus, then continually pretrained on French and German corpora, each for 100 billion tokens. In a second setting, we extend this to five tasks—English, French, German, Arabic, and Japanese—each comprising 100 billion tokens, to evaluate continual pre-training dynamics across a more diverse set of languages. 
We evaluate the models on both retention of earlier knowledge and overall ability on a broad knowledge benchmark. Our study uses the Spectra open LLM suite \citep{kaushal2024spectrasurprisingeffectivenesspretraining}, which provides a range of model sizes within the Llama family of architectures, allowing us to analyze how model scale impacts continual learning dynamics. Concretely, we make the following key contributions to the community:

\begin{figure}[t]
  \centering
  \includegraphics[width=0.9\linewidth]{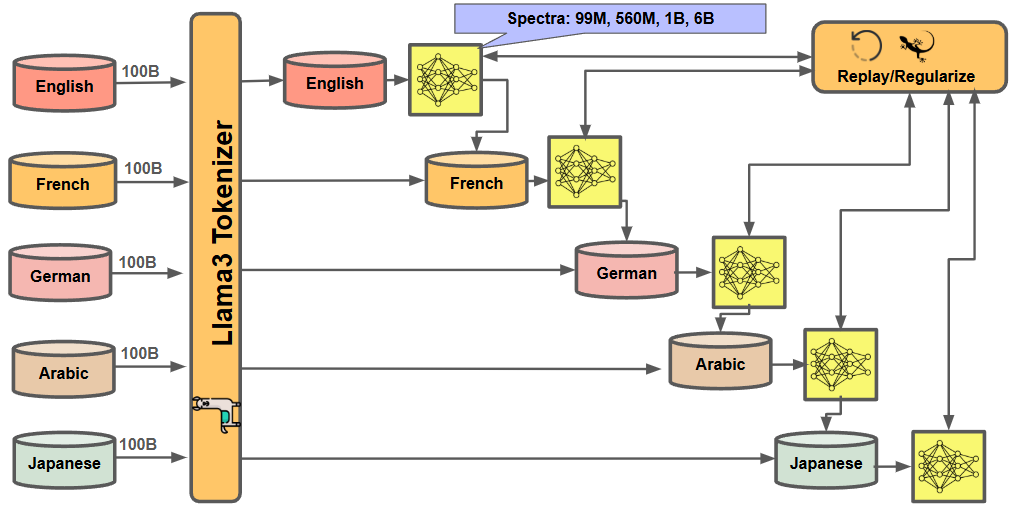}
  \caption{Continual pre-training of Llama models on English, French, German, Arabic and Japanese sequentially using meta-experience replay, which combines replay with gradient alignment through Reptile meta-optimization.}
  \label{fig:cpt_workflow}
\end{figure}
\vspace{-0.2cm}

\begin{itemize}
   \item \textbf{Promoting Gradient Alignment in LLMs.} In the meta-experience replay approach of \citet{riemer2019learninglearnforgettingmaximizing}, the use of Reptile-based meta-learning was motivated as a computationally efficient way of promoting gradient alignment during deep continual learning. However, to the best of our knowledge, gradient alignment approaches for continual learning have not been evaluated in the context of modern LLM models. We are the first to establish that these approaches still lead to benefits in this scaled up setting and provide an updated practical implementation of meta-experience replay geared towards maximizing computational efficiency.
   \item \textbf{More Compute, More Benefits.} Our empirical analysis consistently reveals the interesting insight that providing more compute for replay by using old data as an increasing proportion of each incoming batch leads to more stable learning for both experience replay and meta-experience replay. Indeed, we find that this tradeoff could sometimes be preferable to using larger models that leverage more compute at inference time. Our scaling analysis reveals that adding a 25\% replay rate is more efficient than investing in model size, but that a 50\% replay rate does not further increase performance as efficiently as investing in model size would. 

\end{itemize}


\section{Related Work}
\label{related}

The goal of Continual Pre-Training (CPT) for LLMs is to perform incremental training of pre-existing models on new datasets to enable an expansion of knowledge while still preserving the LLM's original competencies. We refer readers to pre-existing reviews on the landscape of continual learning approaches for an in-depth overview \citep{de2021continual,khetarpal2022towards,shi2024continual}. Here we will just focus on approaches most related to our study. 

\textbf{Replay Buffer Maintenance.} There has been considerable focus on efficient management of replay buffers in the continual learning literature. For example, in RL it is typical to use a recency based FIFO strategy for maintaining a buffer \citep{dqn} because old data is assumed to be outdated (based on old policies). However, in continual supervised learning, where the agents performance is not assumed to influence the data distribution,  it is more common to use reservoir sampling, which has the appealing property of sampling uniformly over past data points encountered \citep{RS,riemer2019learninglearnforgettingmaximizing}. Moreover, when the buffer memory capacity is truly tiny, prior studies have found benefits in further considering clustering \citep{chaudhry2019tinyepisodicmemoriescontinual}, prototypes \citep{icarl} or even scalable memory efficient approximate buffers \citep{GenerativeDist,riemer2019scalable,bashivan2019continual} to make the most use of the available storage. When considering the CPT setting experienced with modern LLMs, it is typical that disk-space is plentiful, but that RAM is severely limited (especially on GPUs). As such, in this work we focus on efficiently storing and loading replay examples on disk rather than maintaining a limited number in memory.   




\textbf{Modular Architectures.} Architectures that exploit modular structure such as sparse mixtures of experts (MoE) or routing models have established benefits in the field of continual learning \citep{Hinton91,Miikkulainen1993,Jordan94,davis2013low,largeneuralnets,Pathnet}. As discussed by \citet{rosenbaum2019routing}, modular architectures with dynamic composition \citep{rosenbaum2018routing,cases2019recursive,chang2018automatically,rosenbaum2019dispatched,thomas,kostas20a,zini2020coagent} have the ability to effect the dynamics of transfer and forgetting by allowing the model to directly orthogonalize weight updates when routing experiences to different modules. \cite{gururangan2021demix} introduced a domain-adaptive continual pre-training approach based on a growing MoE architecture. \cite{jin2021lifelong} extended this idea by continually pre-training RoBERTa-base over a sequence of domain-incremental research paper streams and chronologically ordered tweet streams. Other approaches have sought to mitigate catastrophic forgetting while adapting LLMs to evolving data distributions. \cite{fernandez2024gradientlocalizationimproveslifelong} proposed layer-specific pre-training and adaptive learning rates to preserve knowledge when continually pre-training a GPT2-L model on temporally changing Wikipedia snapshots. \cite{chen2023lifelong} explored lifelong learning from a sequence of online pre-training corpus distributions using a progressively growing MoE architecture. Moreover, recent work by \citet{therien2025continual} explored the influence of the MoE router in effecting the dynamics of continual pre-training. As considerable work has already been focused on modular architectures, and the benefits are largely considered unrelated to replay, we omitted these algorithms in our scaling experiments. 


\textbf{CPT over Multilingual Data.} Several studies have also explored continual pre-training for multilingual adaptation. \cite{duwal2024domainadaptativecontinuallearninglowresource} investigated the adaptation of the Llama3-8B model to Nepali using quantized low-rank adaptation techniques. \cite{gogoulou2024continuallearninglanguageshift} extended this work to three morphologically similar North European languages, scaling models up to 1.3B parameters. However, a comprehensive analysis of forgetting and the use of popular techniques used in continual learning has not been explored in continually pre-training LLMs across diverse scales. In this paper, we dive deeper into this direction. Specifically, we consider the popular "one-pass" setting from the literature in which each task is only visited once. As highlighted by \citep{polynomial}, this setting is very difficult as it is impossible to achieve high frequency updates without introducing considerable myopic optimization bias -- a problem that gets even worse when the Transformer context length is increased \citep{riemerbalancing}.

\section{Our Method: Efficient Meta-Experience Replay (MER)}

In this section, we detail the efficient Meta-Experience Replay (MER) implementation that we consider in this work. First, we discuss the details of our implementation of replay itself. We will then go on to discuss the implementation of MER that is achieved by integrating this replay mechanism with Reptile-based meta-optimization. 

\subsection{Experience Replay}

\textbf{Efficient Replay.} Experience replay seeks to address the stability-plasticity dilemma by maintaining a memory buffer $M$ that is updated at each step $M_{t+1} \gets M_t  \cup \{x\}$. A naive approach that optimizes directly over the memory buffer following the gradient with expected value $\mathbb{E}_{x \sim  M_t} \nabla_{\theta_t} L(x)$ converges to the true unbiased gradient of the long-term steady-state distribution of data points if $p(x|t)$ follows a Markov chain.\footnote{See \citep{khetarpal2022towards} for a discussion of the more general case that arises in RL and continual RL in which the non-stationarity also depends on the behavior of the model itself.} Despite this theoretically appealing property, optimizing directly over the distribution of data in $M$ is a very computationally inefficient way to integrate new data. To see why, consider that the probability of sampling any data point in $M$ is $1/|M|$, implying that the probability of training on any new data point goes down as $|M|$ grows. In this paper, we are interested in exploring practical strategies at scale with a bounded amount of computation per step. As such, in Algorithm \ref{ER} we outline a scalable alternative with a controllable amount of compute overhead, which corresponds to the weighting of past experiences in the stability-plasticity tradeoff. During training, each training step involves a mixed batch. A fraction, $\alpha$, of the batch is filled with randomly drawn samples from $M$ and the rest of the batch is filled with incoming data directly from $p(x|t)$. We consider $\alpha \in \{0, 0.25, 0.5\}$ (0\%, 25\%, and 50\% replay) in our experiments. Note that 50\% replay corresponds to $2 \times$ more FLOPs per gradient step than 0\% replay if the model size $|\theta|$ is held constant. 

\textbf{Buffers Stored on Disk.} In this work, we employ an "infinite" replay buffer to store training examples from previous tasks. Unlike traditional replay buffers that store experiences in RAM, our implementation uses disk storage to maintain a potentially infinite number of samples, limited only by available disk space. To enable scalable and memory-efficient replay in continual pre-training, our disk-backed replay buffer is implemented with asynchronous prefetching and caching, fully compatible with the Megatron/NeoX framework (\cite{gpt-neox-library}). Unlike traditional in-memory buffers, our system splits the total token budget across multiple on-disk files and uses a background thread to populate a cache-aware prefetch queue. This design ensures low-latency access to replay samples without exhausting GPU or host RAM. Samples are stored in raw byte format with offset-based indexing, and metadata tracking allows for full state persistence across training runs. The prefetch queue caches data in memory after it is read from disk, ensuring that batches are readily available when needed and reducing the impact of I/O latency during training. This mechanism is particularly suited for large-scale language model pre-training, where maintaining a diverse replay pool is critical but memory is limited. Our implementation also supports random overwriting when capacity is reached, enabling continuous replacement of samples. Full pseudocode is provided in Algorithm \ref{alg:replay_buffer}, and further implementation details can be found in the repository: \url{https://github.com/chandar-lab/continual-pretraining}.

\subsection{Meta-Experience Replay}

\textbf{Objective.} Following the meta-experience replay approach of \citet{riemer2019learninglearnforgettingmaximizing}, we implement gradient alignment leveraging the first order gradient-based meta-learning approach called Reptile \citep{nichol2018firstordermetalearningalgorithms} (or "Lookahead" \citep{zhang2019lookahead}). Reptile involves a very simple and computationally efficient procedure in which parameters are updated based on a simple interpolation with past parameters every $k$ batches. \citet{nichol2018firstordermetalearningalgorithms} established the surprising theoretical result in their original paper that Reptile approximately optimizes for the following objective over a set of $k$ batches $B_1, ..., B_k$ where $\beta$ is the learning rate (which is assumed to be small): 
\vspace{-0.2cm}
\begin{equation} \label{Reptile}
\argmin_{\theta_t} \mathbb{E}_{B_1, ..., B_k}[2 \sum_{i=1}^k [L(B_i) - \sum_{j=1}^{i-1} \beta \frac{\partial L(B_i)}{\partial \theta_t} \cdot \frac{\partial L(B_j)}{\partial \theta_t}]],
\end{equation}
\textbf{Implementation.} In our implementation of meta-experience replay, the sequence of batches mirrors those from $\alpha$\% replay. We make the small addition of an interpolation every $k$ steps $\theta_t \gets \theta_{t-k} + \epsilon (\theta_{t} - \theta_{t-k})$ where $\epsilon$ is the learning rate of the Reptile meta-update. In our experiments, we set $k=500$ and $\epsilon=0.1$. We have simplified the discussion of the update rule for clarity in the main text. See a detailed description of this procedure in Algorithm \ref{MER}. Note that the addition of Reptile updates adds negligible overhead, both in terms of compute and memory requirements. By optimizing for this regularized objective, as demonstrated in \citep{riemer2019learninglearnforgettingmaximizing}, Reptile enables the model to learn parameters that promote transfer and avoid interference across updates to the extent that the model can generalize. We consider a fairly large value of $k$ in order to promote stronger regularization in our experiments. By learning to align gradients, meta-experience replay can make more efficient use of its compute spent on data from the replay buffer if the model can learn general knowledge about the nature of old data.  
\section{Experimental Setup}
\label{methodology}
Our MER continual pre-training approach augments standard LLM training with two components: a replay buffer that stores samples from previous datasets, and gradient alignment induced by a Reptile-based meta-update applied periodically after a number of training batches. This section details the model architecture we leverage, the specific sequential training process we explore, and how replay and Reptile are implemented in our experiments.

\subsection{Model Architecture}
In this work, we adopt the Spectra LLM suite \citep{kaushal2024spectrasurprisingeffectivenesspretraining} as our base models which uses the llama architecture. Spectra models are Transformer-based autoregressive language models of varying sizes. In particular, we use four model variants from the Spectra suite: 99M, 560M, 1B and 6B parameters. Using a consistent model family throughout our experiments allows us to attribute differences in performance solely to model scale rather than architecture.\footnote{We denote the 1.1B Spectra model as 1B, and the 5.7B Spectra model as 6B for brevity throughout this paper.}

\subsection{Continual Pre-Training Process}
The training sequence has three stages (as shown in Figure \ref{fig:cpt_workflow}):
For Task A, we use a subset of the DCLM baseline dataset provided by DataComp-LM \citep{li2024datacomplmsearchgenerationtraining}: this is a high-quality English text corpus derived from the CommonCrawl. We sampled 100B tokens out of the 240T token standardized corpus. Tasks B and C are French and German respectively drawn from a subset of the 166 language Open Super-large Crawled Aggregated coRpus also known as the OSCAR Dataset (unlabeled CommonCrawl extractions) \citep{suarez2019asynchronous,suarez2020monolingual,abadji2022cleanerdocumentorientedmultilingualcrawled} with 100B tokens each, to represent learning new languages. In an extended setting, we evaluate continual pre-training across five language tasks: English, French, German, Arabic, and Japanese, each with 100 billion tokens. The Arabic data is sourced from \cite{aloui2024101billionarabicwords}, while the Japanese data is constructed from a large-scale CommonCrawl-derived corpus \cite{abeja_cc_ja_techblog2024}. We deliberately choose distinct languages to create a challenging continual learning scenario. After the initial pre-training phase on English, the model must learn French as a different distribution and then German, without forgetting English. In the extended five-task setting, Arabic and Japanese further increase the distributional diversity, requiring the model to adapt to significantly different linguistic structures while retaining performance on previously learned languages. There is minimal overlap in content between these datasets, and each is sufficiently large to train new knowledge, thus posing a high risk of catastrophic forgetting of earlier languages if no countermeasures are applied.
Table \ref{tab:dataset_composition} details the dataset composition used in our setup. We use a consistent training recipe across tasks: each 100B token task is processed for a fixed number of steps. We use the AdamW optimizer with a learning rate following established experiments for the spectra model suite \citep{kaushal2024spectrasurprisingeffectivenesspretraining}. We use a cosine learning rate schedule  with a linear warmup of 357 steps as shown in Figure \ref{fig:learning_rate} and a cosine decay phase after that along with an effective batch size of 4096. We train for one epoch over each task’s data ensuring the model sees 100B tokens per stage. There is no gradient or optimizer resetting between tasks – the training seamlessly continues.

\subsection{Evaluation Metrics}

We consider the following evaluation metrics for assessing continual pre-training performance in LLMs: 


\paragraph{Forgetting Score:} The degradation in performance on previously encountered tasks as new tasks are introduced. This is critical in continual pre-training, where catastrophic forgetting can impair long-term knowledge retention. We measure forgetting using the increase in validation loss relative to the best past performance achieved on that task:

\vspace{-2mm}
\begin{equation*}
    \text{forgetting score}_{\text{current task}} = 
  \min_{\text{past}}\text{validation loss} - \text{validation loss}_{\text{current task}}
\end{equation*}
\vspace{-4mm}

\paragraph{Retained Loss:} The average cross-entropy loss on the validation set of all tasks at the end of training. This metric is similar to forgetting score in that it is a measure of the stability of the model, but it is better suited to scaling analysis as it is constrained to positive values. We use forgetting score in our other experiments as it is more interpretable. 

\paragraph{Learned Loss:} The average cross-entropy loss on the validation set of each task directly after training on that task. This metric focuses instead on the plasticity of the model in that it measures the average quality of the model's ability to adapt to new tasks. Notice that loss of performance due to forgetting does not influence this metric at all. 


\paragraph{Downstream Performance:} We evaluate generalization ability by testing on a range of downstream benchmarks (e.g., HellaSwag \citep{zellers2019hellaswagmachinereallyfinish}, PiQA \citep{bisk2019piqareasoningphysicalcommonsense}, and PubMedQA \citep{jin2019pubmedqadatasetbiomedicalresearch}), which provide insight into the LLM's ability to generalize to new tasks and domains despite not being specifically trained on them. 

\subsection{CPT Methods and Baselines}
\label{setup}




We conduct experiments for each Spectra model size (99M, 560M, 1B and 6B parameters) under a range of settings:
\begin{itemize}
    \item \textbf{Sequential Baseline (No Replay, No Reptile):} Simple sequential training on A → B → C with no special measures. This is expected to highlight catastrophic forgetting, especially on A (English) after tasks B and C.
    \item \textbf{Replay only:} CPT with a replay buffer and without any gradient alignment i.e., no Reptile based updates ($\epsilon=0$). We test replay ratios $\alpha = \{25\%, 50\% \}$ to assess different levels of replay compute investment.
    \item \textbf{Reptile only:} Continual training with only gradient alignment but no replay. This isolates the effect of Reptile-based updates without any stored data.
    \item \textbf{Replay + Reptile (MER):} Our full method, combining the replay buffer with replay ratios $\alpha = \{25\%, 50\% \}$ along with gradient alignment induced by Reptile-based meta-updates.
    \item \textbf{Joint Training Baseline:} Model trained on the joint A+B+C dataset in an i.i.d. setting together from scratch. 
\end{itemize}

\begin{figure}[h!]
  \centering
  \begin{subfigure}[b]{0.48\textwidth}
    \centering
    \includegraphics[width=\linewidth, height=5cm]{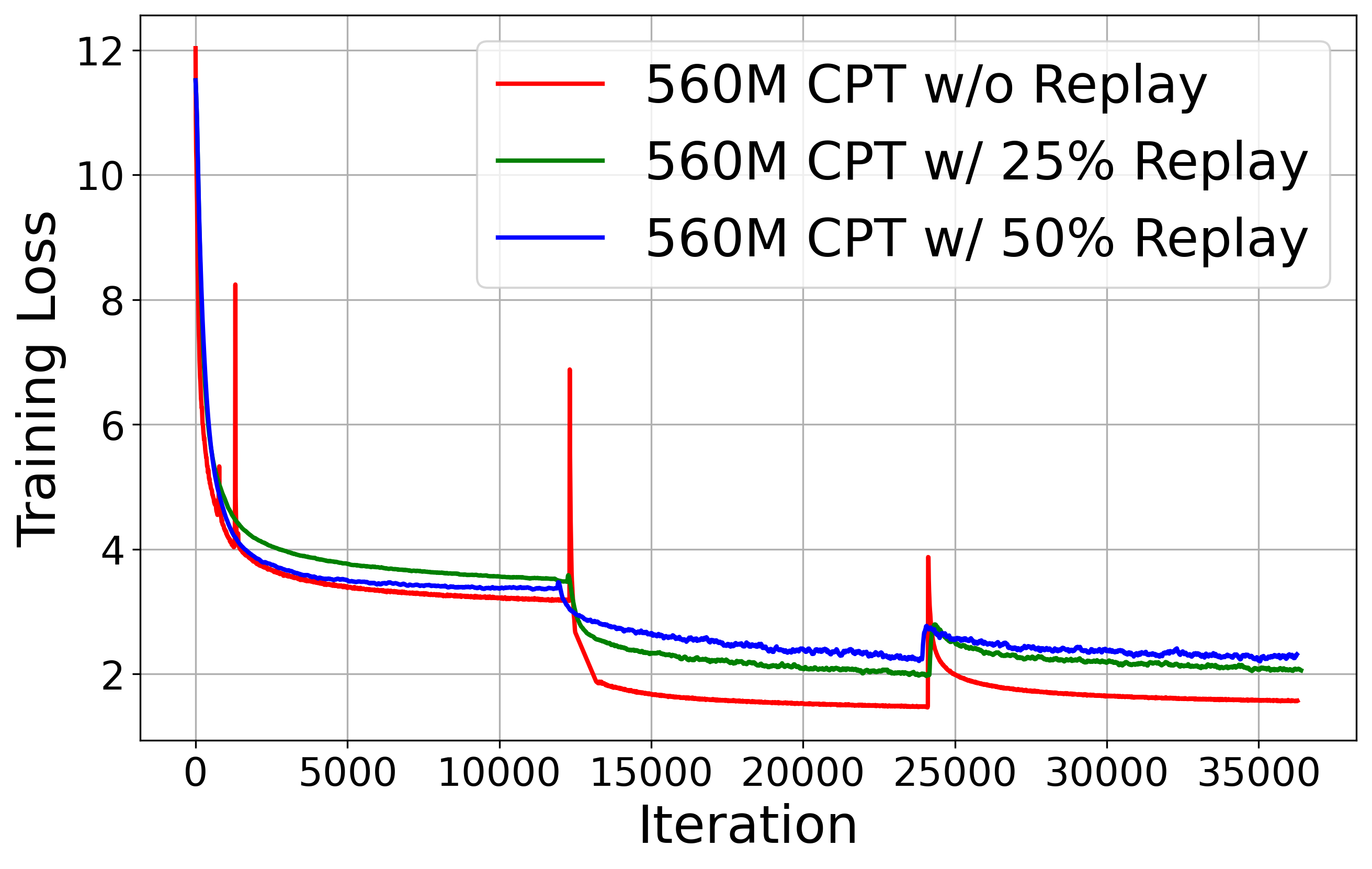}
    \caption{Training loss on current task}
    \label{fig:train_loss_curve}
  \end{subfigure}%
  \hfill
  \begin{subfigure}[b]{0.48\textwidth}
    \centering   
    \includegraphics[width=\linewidth, height=5cm]{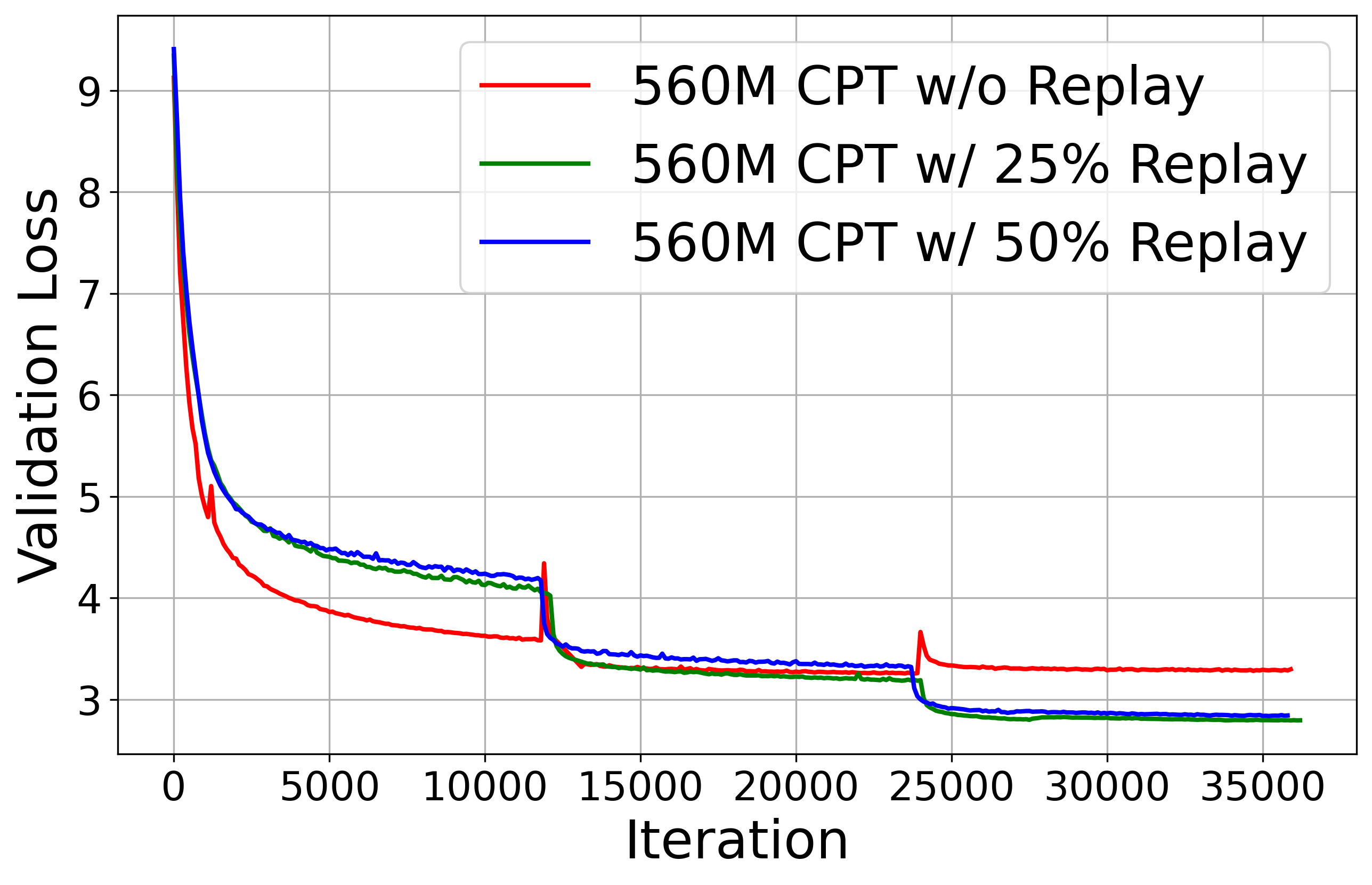}
    \caption{Validation loss on union of validation sets across tasks}
    \label{fig:val_loss_curve}
  \end{subfigure}
  \caption{Cross-entropy loss curves for the 560M parameter Spectra model with varying levels of replay. Note: In Figure \ref{fig:val_loss_curve} the validation set is a union of the validation sets from all three tasks.}
  \label{fig:loss_curves_560M}
\end{figure}

\section{Results}
\label{others}

We conducted a series of experiments to address several key questions about CPT in this section: 
\begin{itemize}
    \item \textbf{Q1: Impact of Experience Replay in CPT:} What is the effect of experience replay and how does varying replay proportions affect the model's stability and plasticity across model sizes?
    \item \textbf{Q2: Gradient Alignment \& Replay Synergy:} Does gradient alignment synergize with experience replay to reduce forgetting, promote plasticity, and improve downstream performance?
    \item \textbf{Q3: Generalization via Alignment:} Does combining gradient alignment with replay lead to improved generalization across tasks, as evidenced by validation loss trends across tasks?
    \item \textbf{Q4: Computational Scaling of Replay \& Gradient Alignment:} When given a budget of FLOPs per incoming token is it wiser to invest resources in increased replay, gradient alignment, or increased model size?
\end{itemize}

\subsection{Impact of Experience Replay in CPT}
\label{sec:replay_impact_cpt}

Experience replay helps neural networks remember old information, but this requires more computation. Our experiments compare models that use replay to those that simply use more parameters to achieve the same computational cost. We found strong evidence that replay can be a more efficient way to maintain performance.

\begin{figure}[h]
    \centering
    \includegraphics[width=0.4\textwidth]{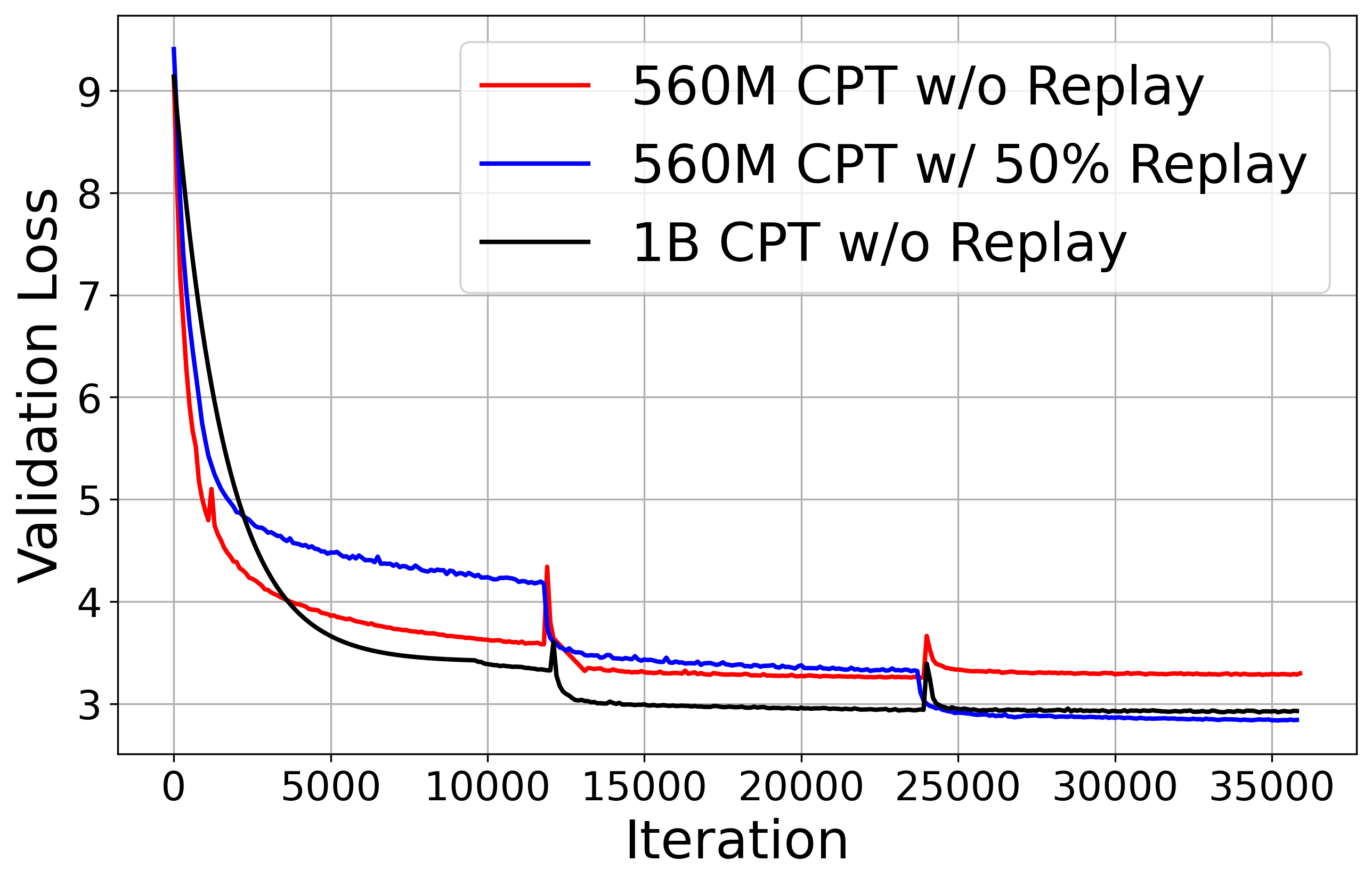}
    \caption{Comparison of cross-entropy validation loss curves for the Spectra 560M model with replay to the Spectra 1B model without replay. Note: In Figure \ref{fig:validation_comparaison} the validation set is a union of the validation sets from all three tasks.}
    \label{fig:validation_comparaison}
\end{figure}

Figure \ref{fig:loss_curves_560M} presents training and validation loss curves for a 560M parameter model continually pretrained with varying levels of replay (0\%, 25\%, and 50\%). As illustrated in Figure \ref{fig:val_loss_curve}, the 560M model trained without replay (red line) shows the worst performance, with a validation loss reaching approximately 3.3 at the end of training, which means it forgot a lot. In contrast, the 560M model with 50\% replay (blue line) performed the best, with a validation loss stabilizing around 3.0. The 560M model with 25\% replay (green line) had intermediate results. This clear trend shows that more replay helps the model remember what it learned earlier, leading to better overall performance. Figure \ref{fig:validation_comparaison} directly compares the 560M model with replay to a 1B model trained without replay. While the 1B model achieves a validation loss close to the 560M model with 50\% replay, the 560M model needs way fewer parameters. Because the 1B model has nearly twice the parameters of the 560M parameter model, this shows that replay can be more efficient at retaining old knowledge. Experience replay appears to be a more efficient way to continually pretrain models, giving similar results to a bigger model but with smaller inference costs. We observe similar training and validation curves for the experiment conducted on a large-scale model with 6B parameters, further supporting the scalability of replay.

\subsection{Synergy of Gradient Alignment and Replay}
\label{sec:replay_reptile_synergy}
This section shows how gradient alignment, implemented via Reptile-style updates, complements experience replay to further mitigate forgetting and improve downstream performance.

\begin{figure}[h]
    \centering
    \begin{minipage}{\textwidth}
        \centering
        \includegraphics[width=0.9\textwidth]{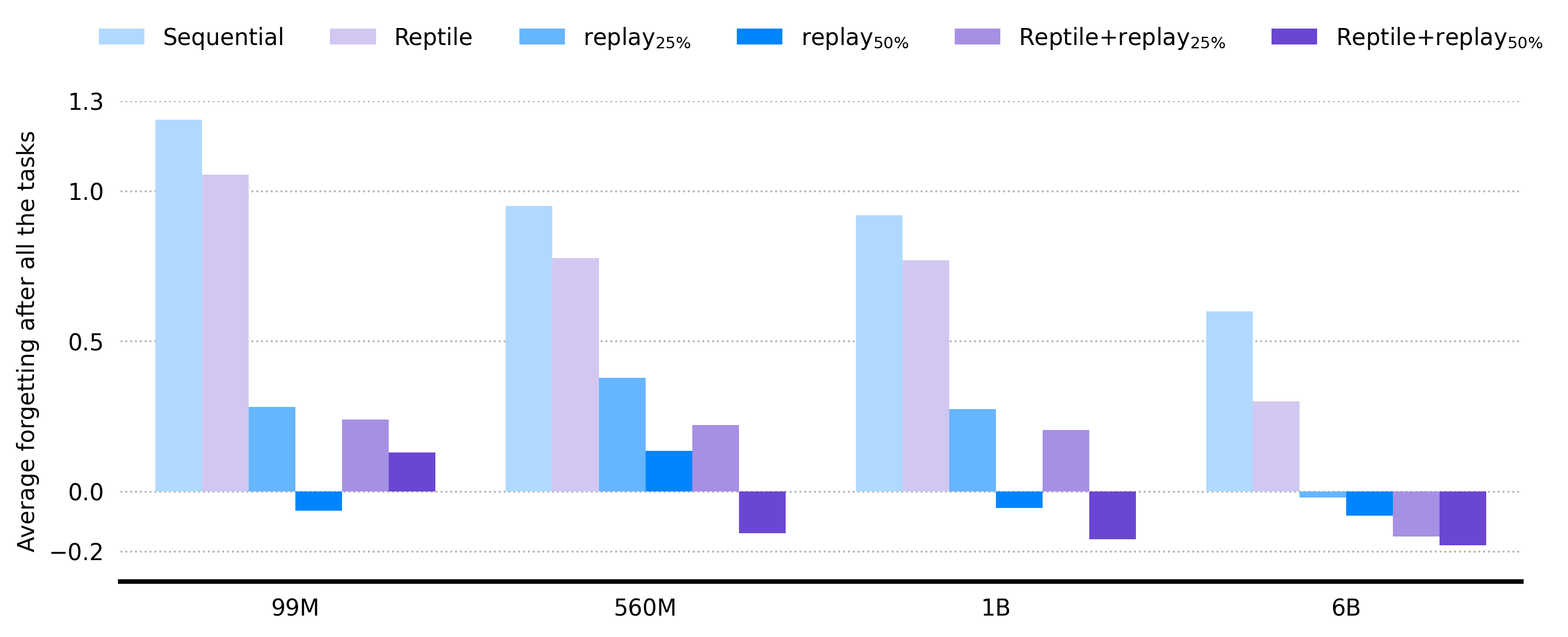}
        \subcaption{{Final average forgetting score for different model configurations across four model sizes: 99M, 560M, 1B, and 6B parameters}}
        \label{fig:forgetting_bar_plot_a}
    \end{minipage}
    
    \vspace{1em} 

    \begin{minipage}{\textwidth}
        \centering
        \includegraphics[width=0.8\textwidth]{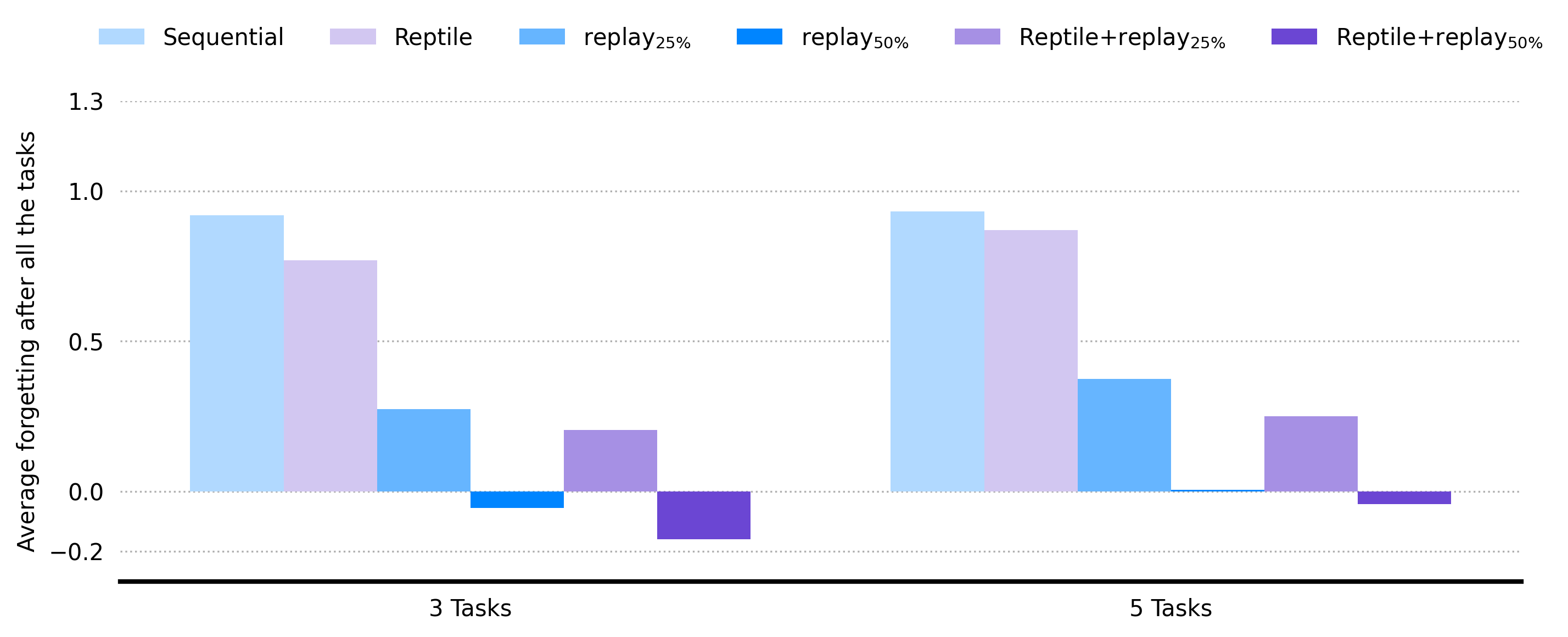}
        \subcaption{{Final average forgetting score for different configurations of the 1B model while varying the number of tasks from 3 to 5}}
        \label{fig:forgetting_bar_plot_b}
    \end{minipage}
    \caption{{Final Average Forgetting Scores Across Scale}}
    \label{fig:forgetting_plots}
\end{figure}

This reduction is quantified in the forgetting bar plot in Figure \ref{fig:forgetting_bar_plot_a}. Here, we observe a consistent trend across all model sizes: 50\% replay with Reptile consistently achieves the lowest average forgetting score. Figure \ref{fig:forgetting_bar_plot_a} reveals a significant decrease in forgetting for 50\% replay with Reptile compared to replay or Reptile alone. We have also extended our analysis to a larger model with 6B parameters, where the effectiveness of the 50\% replay with Reptile in mitigating forgetting remains consistent. Furthermore, results from the pre-training experiments with the 1B parameter model—extended from 3 tasks to 5 tasks—show similar trends (see \ref{fig:forgetting_bar_plot_b}), reinforcing the robustness of our approach under increased task scaling. As the number of tasks increases, most methods exhibit higher average forgetting, highlighting the scalability challenge in continual learning. While sequential and Reptile-only training degrade significantly with more tasks, replay-based strategies, particularly 50\% replay with Reptile, maintain strong performance, even achieving negative forgetting, which suggests enhanced retention or backward transfer. This indicates their suitability for longer, more complex continual learning sequences.

Table \ref{tab:model_performance} provides a quantitative assessment of downstream task performance on Hellaswag, Piqa, and Pubmedqa. For the 560M parameter models, adding 25\% replay, while reducing forgetting, achieves downstream task performance comparable to 560M joint model. However, Table \ref{tab:model_performance} highlights a crucial synergy: combining replay with Reptile significantly boosts downstream task performance. Specifically, the 560M model using 25\% replay and Reptile achieves an average score of 67.5, outperforming the 560M model trained jointly (67.33), as well as the 560M model with only replay (66.4) and the 560M model with only Reptile (65.4). This strongly suggests that gradient alignment and replay provide complementary benefits, allowing the model to both retain previously learned knowledge and generalize effectively to new tasks. The trend observed with the 560M models is also apparent in the 99M models (Table \ref{tab:model_performance}), although the absolute performance gains are smaller. Conversely, the 1B models, likely due to their increased capacity and already strong performance, do not show a substantial average improvement from the combined approach. However, examining the individual task scores for the 1B models reveals nuanced benefits. For the 1B models, both replay and Reptile individually improve the Hellaswag, Piqa, and Pubmedqa scores. For example, the Reptile gives the best Pubmedqa performance. This supports the notion that even for larger models, gradient alignment and replay can contribute to improved performance on specific downstream tasks.
The benefits are most evident in the 6B models. The joint training baseline average is only 72.6, while 25\% replay with Reptile yields 76.8, and 50\% replay with Reptile achieves the best overall result of 77.1. This is quite unexpected and suggests that strong CPT models can benefit from a form of curriculum learning once they overcome optimization instability. These results confirm that even at high model capacities, combining replay with gradient alignment yields consistent and measurable improvements.

\subsection{Generalization via Alignment}
\label{sec:generalization_alignment}
\begin{figure}[h]
    \centering
    \includegraphics[width=0.9\textwidth]{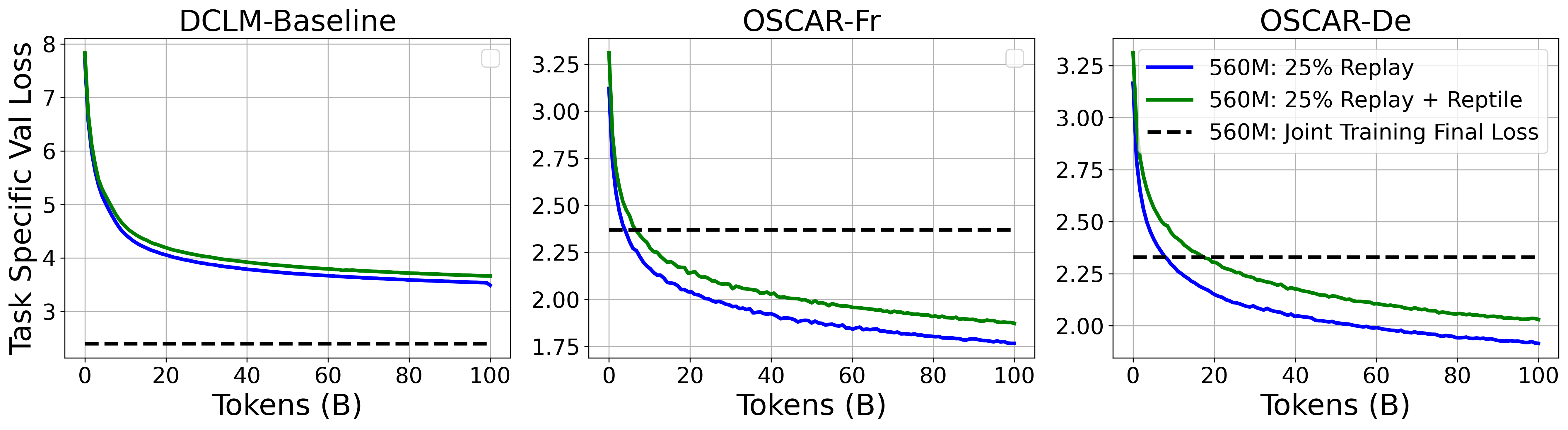}
    \caption{Task-specific cross-entropy validation loss curves for a 560M model during CPT where we evaluate 25\% replay and 25\% replay with Reptile against the final loss of the joint training baseline.}
    \label{fig:task_val_loss}
\end{figure}

In this section, we show that gradient alignment is a key attribute for promoting generalization across different tasks during continual pre-training.
As brought up in Section \ref{sec:replay_reptile_synergy}, we discuss such generalization results in more detail here. Figure \ref{fig:task_val_loss} displays task-specific validation loss curves for the 560M models, offering direct visual evidence of Reptile's impact on generalization. For both OSCAR-Fr and OSCAR-De, the addition of Reptile (green) pulls the validation loss curve much closer to the joint training baseline (black dotted line) compared to the model with only 25\% replay (blue). This demonstrates Reptile's ability to balance learning new knowledge with minimal forgetting.


Table \ref{tab:dclm_french_german} also presents a quantitative confirmation of gradient alignment's impact on generalization. Through the average validation loss values at the end of training in the table we observe that Reptile-style updates serve as a critical attribute for enhancing generalization during continual pre-training.

\subsection{Computational Scaling Analysis}

\begin{figure}[h!]
    \centering
    \includegraphics[width=0.8\linewidth]{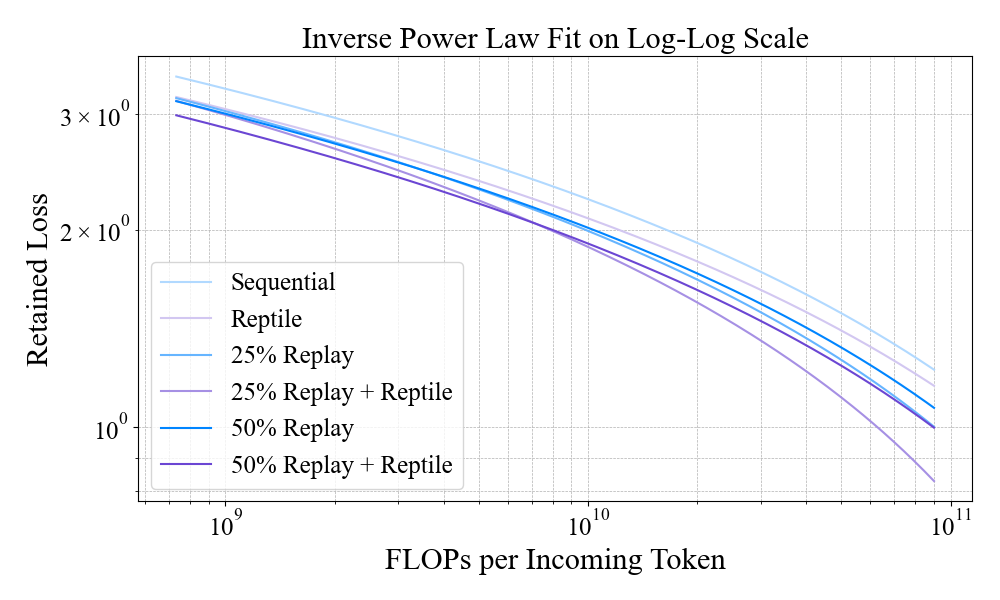}
    \caption{\textbf{Stability Scaling Analysis: Retained Loss vs. Compute per Token}. We consider the average retained loss across tasks at the end of training for each model and plot the results of an inverse power law fit for each model family. }
    \label{fig:scaling_plot_compute}
\end{figure}

\begin{figure}[h!]
    \centering
    \includegraphics[width=0.8\linewidth]{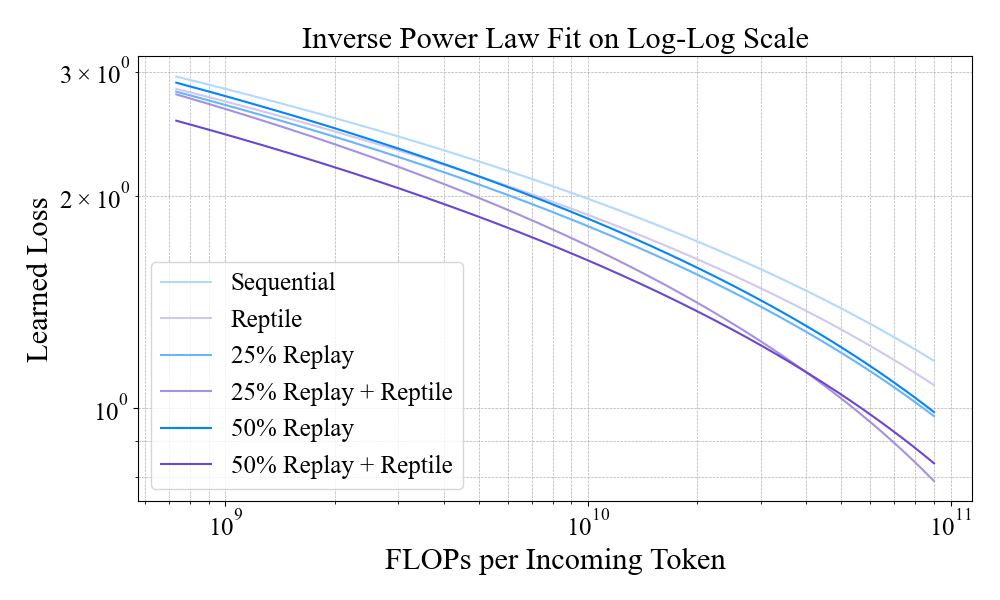}
    \caption{\textbf{Plasticity Scaling Analysis: Learned Loss vs. Compute per Token}. We consider the average learned loss across tasks during training for each model and plot the results of an inverse power law fit for each model family. }
    \label{fig:scaling_plot_compute_learned}
\end{figure}

We provide our main scaling analysis in Figures \ref{fig:scaling_plot_compute} and \ref{fig:scaling_plot_compute_learned}. In Figure \ref{fig:scaling_plot_compute} we provide an analysis of the stability of each model through measurements of the retained performance across tasks at the end of training. Meanwhile, in Figure \ref{fig:scaling_plot_compute_learned} we provide an analysis of the plasticity of each model through measurements of the performance on each task directly once training on it is complete. What we plot here are inverse power law best fit lines of the raw data provided in Table \ref{tab:dclm_french_german} and Table \ref{tab:plasticity_tab}. These results are further contextualized by the results in Figures \ref{fig:scaling_plot_model_size} and \ref{fig:scaling_plot_model_size_learned} for scaling with respect to model size, demonstrating that models of the same size consistently get better stability and plasticity performance with more replay than they do with less. However, Figures \ref{fig:scaling_plot_compute} and \ref{fig:scaling_plot_compute_learned} highlight that it would be better to invest compute in growing the model size for large models with 25\% replay than it is to invest the compute in increasing the replay rate. This can be somewhat expected based on the fact that 25\% replay leads to a $1.33\times$ increase in FLOPs per token over the sequential model whereas 50\% replay leads to a $2\times$ increase. One of the remarkable things about the results in Figures \ref{fig:scaling_plot_compute} and \ref{fig:scaling_plot_compute_learned} are the nearly free gains in both stability and plasticity performance that come from the integration of Reptile. In our experiments, Reptile only added a computational overhead of three times the model size in FLOPs every 500 batches, which is negligible relative to the cost of gradient updates. If anything, it seems like 25\% replay with Reptile integrated is becoming better with model size even faster than 25\% replay without Reptile. This is preliminary evidence of the very exciting possibility that improvements at small scales related to meta-learning can lead to even bigger improvements than those experienced by vanilla learning with large scales of compute.

\section{Discussion}
Our experiments demonstrate that experience replay effectively mitigates forgetting in LLM pre-training while offering a resource-efficient alternative to model scaling. 
Our results also provide strong evidence that gradient alignment, implemented here via Reptile, and experience replay are not mutually exclusive but rather synergistic approaches for continual pre-training. The combination of both leads to a MER model that not only better retains previously learned knowledge but also demonstrates enhanced plasticity, and generalizes more effectively to downstream tasks. 

\paragraph{Limitations:} Our study is largely limited to three tasks in a fixed sequence. Real-world continual learning for LLMs might involve many more stages, possibly with revisiting domains. To explore this, we also evaluate one model scale on a five-task sequence and observe that the synergistic benefits of our approach persist, suggesting its potential scalability. However, further validation is needed. 
Another limitation is evaluation scope: a deeper evaluation such as directly measuring knowledge changes, or testing multilingual QA throughout learning would provide greater insight about the evolution of factual knowledge, which is largely opaque based on our experiments.


\section{Conclusion}


In this work, we demonstrate for the first time that experience replay and gradient alignment, implemented through Reptile-style updates, are synergistic and compute efficient techniques for continual pre-training of large language models. 
Although larger models inherently benefit from increased scale, replay remains valuable, allowing for a more efficient balance between training resources, the model's stability with respect to previous acquired knowledge and the model's plasticity with respect to new knowledge. Moreover, gradient alignment seems to result in a consistent performance boost in both stability and plasticity with negligible compute overhead across model scales. A combined approach, leveraging both replay and gradient alignment, leads to models that not only better retain knowledge but also demonstrate an improved ability to adapt to new tasks and generalize more effectively to downstream tasks. 
While replay remains popular in the literature, we hope our work inspires more focus on the potential additional benefits of gradient alignment as the field looks to build efficient approaches for continual pre-training of foundation models. 

\section{Acknowledgment}
This research was supported in part by the Oak Ridge Leadership Computing Facility at Oak Ridge National Laboratory, through computing resources on the Frontier supercomputer, provided under the ALCC 2024 program award “Scalable Foundation Models for Transferable Generalist AI.” Oak Ridge National Laboratory is supported by the U.S. Department of Energy, Office of Science, under Contract No. DE-AC05-00OR22725. We extend special thanks to Jens Glaser for valuable assistance with the Summit supercomputer. We would also like to thank Darshil Doshi and Quentin Antony for their contributions throughout the project. This work was further supported by the Canada Excellence Research Chairs (CERC) Program, Fujitsu Corporation, and the Canadian Institute for Advanced Research (CIFAR).

\bibliography{collas2025_conference}
\bibliographystyle{collas2025_conference}

\appendix
\section{Appendix}
\subsection{Dataset and Training Setup}
\begin{figure*}[h]
    \centering
    \begin{minipage}{0.45\textwidth} 
        \centering
        \begin{tabular}{|c|c|c|}
            \hline
            \textbf{Dataset} & \textbf{Language} & \textbf{Tokens (B)} \\ \hline
            DCLM-Baseline & English & 100 \\ \hline
            OSCAR-Fr & French & 100 \\ \hline
            OSCAR-De & German & 100 \\ \hline
            ClusterlabAi Dataset  & Arabic & 100\\ \hline
            Abeja-cc-ja & Japanese & 100 \\ \hline
        \end{tabular}
        \captionof{table}{Dataset Composition for Continual Pre-training}
        \label{tab:dataset_composition}
    \end{minipage}
    \begin{minipage}{0.5\textwidth} 
        \centering
        \includegraphics[width=\linewidth]{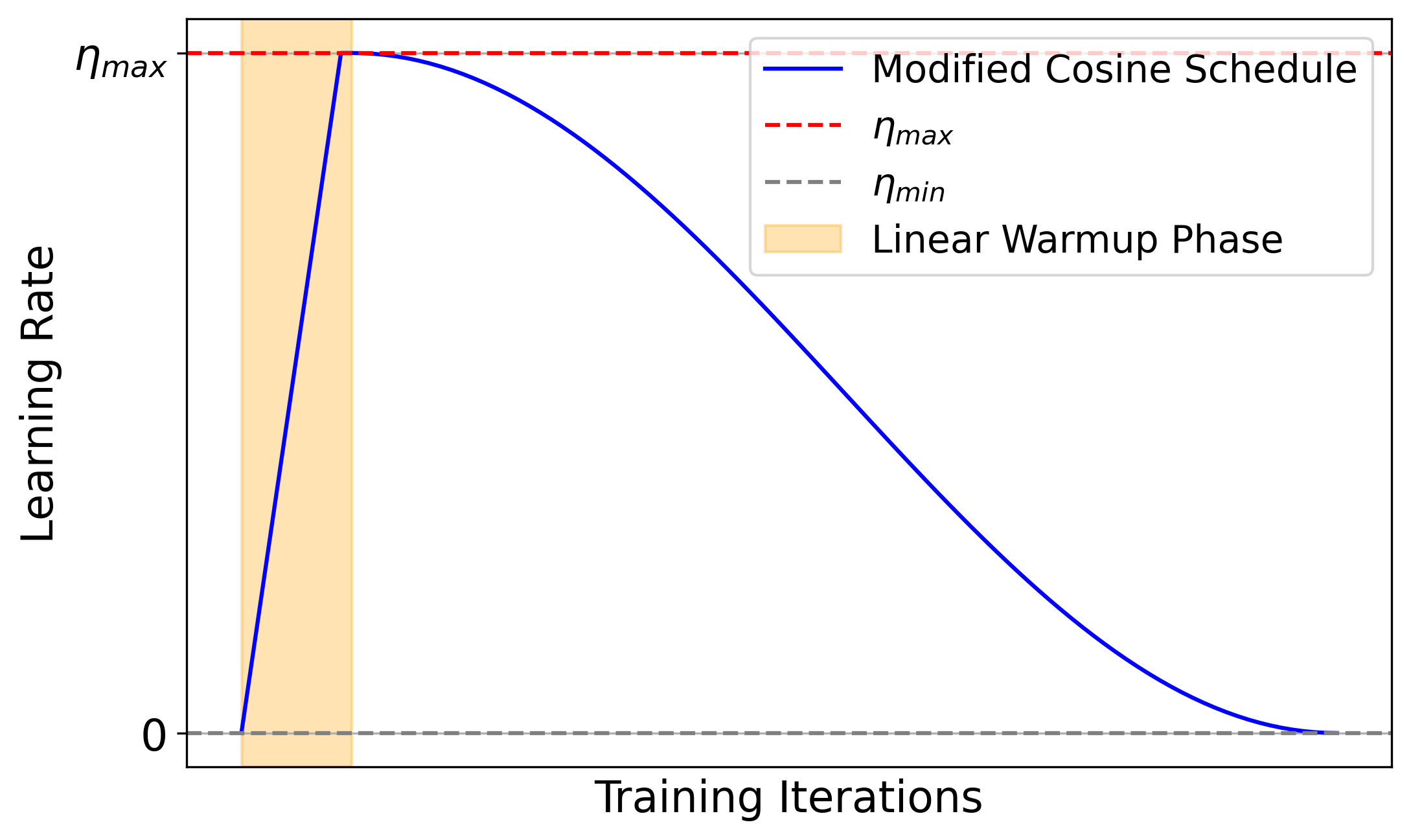} 
        \captionof{figure}{Learning rate schedule}
        \label{fig:learning_rate}
    \end{minipage}
\end{figure*}

To support our continual pre-training experiments, we curated a multilingual corpus with balanced token distribution across five languages. Table~\ref{tab:dataset_composition} summarizes the dataset composition. Each language-specific corpus contains 100 billion tokens, ensuring a comparable scale of training data across tasks. We utilize a cosine learning rate schedule for stable convergence, shown in Figure~\ref{fig:learning_rate}.

\subsection{Retained Performance Across Tasks}

Table~\ref{tab:dclm_french_german} reports the validation loss observed at the end of training across the first three tasks in the continual pre-training sequence (English → French → German). Lower retained loss values indicate better knowledge preservation across previously learned tasks.

\begin{table}[h]
    \centering
    \renewcommand{\arraystretch}{1.2}
    \begin{tabular}{lccc|c}
        \toprule
        \textbf{Training Method} & \multicolumn{3}{c|}{\textbf{Validation Loss}} & \textbf{AVG} \\
        & $\mathcal{D}_0$ DCLM & $\mathcal{D}_1$ French & $\mathcal{D}_2$ German &  \\
        \midrule
        99M (No Replay) & 4.70 & 2.40 & 2.60 & 3.23 \\
        99M (25\% Replay) & 4.03 & 2.28 & 2.27 & 2.86 \\
        99M (50\% Replay) & 3.60 & 2.00 & 2.36 & 2.65 \\
        99M (Reptile Only) & 4.10 & 2.35 & 2.40 & 2.95 \\
        99M (25\% Replay + Reptile) & 3.98 & 2.13 & 2.29 & 2.80 \\
        99M (50\% Replay + Reptile) & 3.55 & 1.87 & 2.01 & 2.47 \\
        \hline
        560M (No Replay) & 4.20 & 2.10 & 2.40 & 2.90 \\
        560M (25\% Replay) & 3.53 & 1.76 & 1.91 & 2.40 \\
        560M (50\% Replay) & 3.12 & 1.42 & 2.34 & 2.29 \\
        560M (Reptile Only) & 4.00 & 2.05 & 2.30 & 2.78 \\
        560M (25\% Replay + Reptile) & 3.38 & 1.68 & 1.98 & 2.35 \\
        560M (50\% Replay + Reptile) & 2.90 & 1.21 & 1.60 & 2.24 \\
        \hline
        1B (No Replay) & 3.70 & 1.80 & 2.00 & 2.50 \\
        1B (25\% Replay) & 2.78 & 1.87 & 2.22 & 2.29 \\
        1B (50\% Replay) & 2.50 & 1.41 & 2.33 & 2.08 \\
        1B (Reptile Only) & 3.00 & 1.99 & 2.12 & 2.37 \\
        1B (25\% Replay + Reptile) & 2.62 & 1.65 & 2.28 & 2.18 \\
        1B (50\% Replay + Reptile) & 2.50 & 1.65 & 1.83 & 1.99 \\
        \hline
        6B (No Replay) & 2.00 & 1.30 & 1.10 & 1.47 \\
        6B (25\% Replay) & 1.30 & 1.05 & 0.90 & 1.08 \\
        6B (50\% Replay) & 1.17 & 0.93 & 0.80 & 0.97 \\
        6B (Reptile Only) & 1.70 & 1.21 & 1.09 & 1.33 \\
        6B (25\% Replay + Reptile) & 1.11 & 0.86 & 0.77 & 0.91 \\
        6B (50\% Replay + Reptile) & 1.05 & 0.80 & 0.74 & 0.86 \\
        \bottomrule
    \end{tabular}
    \caption{Retained validation loss comparison for individual tasks across multiple models during CPT for the DCLM → French → German sequence. The AVG column represents the mean performance across the three datasets.}
    \label{tab:dclm_french_german}
\end{table}



\subsection{Generalization on Downstream Tasks}
Table~\ref{tab:model_performance} compares performance across downstream tasks such as HellaSwag, PiQA, and PubMedQA.

\begin{table}[h]
\centering
\begin{tabular}{|c|c|c|c|c|c|}
\hline
\textbf{Model} & \textbf{Hellaswag (en) } & \textbf{Piqa} & \textbf{Pubmedqa} & \textbf{Average (en)} & \textbf{Hellaswag (fr) } \\
\hline
99M joint & 30.1 & 79.3 & 46.3 & 51.9 & 28.2 \\
99M sequential & 30.1 & 52.3 & 43.3 & 41.9 & 27.5\\
99M w/ 25\% Replay & 31.1 & 71.4 & 49.3 & 50.6 & 27.5 \\
99M w/ 50\% Replay & 40.8 & \textbf{74.3}~$\uparrow$ & 51.2 & 58.8 & 27.8 \\
99M w/ Reptile & 29.9 & 67.2 & 45.3 & 50.8 & 27.6  \\
99M w/ 25\% Replay + Reptile & 42.2 & 72.8 & 48.2 & 61.1 & 29.4 \\
99M w/ 50\% Replay + Reptile & \textbf{42.3}~$\uparrow$ & 73.4 & \textbf{58.9}~$\uparrow$ & \textbf{61.5}~$\uparrow$  & \textbf{30.6}~$\uparrow$ \\
\hline
560M joint & 44.6 & 80.3 & \textbf{57.1}~$\uparrow$ & 60.6 & 42.5 \\
560M sequential & 42.3 & 75.6 & 56.2 & 60.7 & 40.1 \\
560M w/ 25\% Replay & 42.6 & 80.3 & 56.2 & 58.0 & 41.9 \\
560M w/ 50\% Replay & 43.1 & 82.3 & 56.8 & 59.7 & 43.1 \\
560M w/ Reptile & 42.4 & 78.3 & 55.6 & 60.7 & 41.6 \\
560M w/ 25\% Replay + Reptile & 43.0 & 82.7 & 56.7 & 58.8 & 42.6 \\
560M w/ 50\% Replay + Reptile & \textbf{43.5}~$\uparrow$ & \textbf{83.0}~$\uparrow$ & 57.0 & 	\textbf{61.2}~$\uparrow$ & \textbf{43.2}~$\uparrow$ \\
\hline
1B Joint & 44.8 & 82.6 & 59.7 & 62.4 & 41.5 \\
1B sequential & 44.3 & 79.2 & 58.6 & 60.7 & 38.0 \\
1B w/ 25\% Replay & 44.2 & 81.4 & 58.9 & 61.5 & 39.3\\
1B w/ 50\% Replay & \textbf{46.6}~$\uparrow$ & \textbf{84.6}~$\uparrow$ & \textbf{60.7}~$\uparrow$ & \textbf{64.0}~$\uparrow$ & 41.5 \\
1B w/ Reptile & 43.8 & 78.7 & 58.2 & 60.2 & 37.8 \\
1B w/ 25\% Replay + Reptile & 44.6 & 79.6 & 60.2 & 61.5 & 41.3 \\
1B w/ 50\% Replay + Reptile & 46.5 & 84.2 & 60.6 & 63.8 & \textbf{41.8}~$\uparrow$ \\
\hline
6B joint & 67.8 & 86.3 & 73.9 & 76.0 & 55.6\\
6B sequential & 55.3 & 86.1 & 66.5 & 69.3 & 53.2 \\
6B w/ 25\% Replay & 66.7 & 86.8 & 73.2 & 75.5 & 54.7\\
6B w/ 50\% Replay  & 66.9 & 86.9 & 64.9 & 71.0 & 55.4\\
6B w/ Reptile  & 58.1 & 86.5 & 68.6 & 71.1 & 54.3 \\
6B w/ 25\% Replay + Reptile  & 68.6  & 87.1  & 74.8  & 76.8 & 56.1 \\
6B w/ 50\% Replay + Reptile & \textbf{68.9}~$\uparrow$ & \textbf{87.3}~$\uparrow$ & \textbf{75.1}~$\uparrow$ & \textbf{77.1}~$\uparrow$ & \textbf{56.7}~$\uparrow$ \\

\hline
\end{tabular}
\caption{Accuracy comparison of different models across three datasets. The best-performing values in each model size category are highlighted in bold with an upper arrow ($\uparrow$) indicating that higher values are better. The \textbf{Average} column represents the mean performance across the three benchmarks.}
\label{tab:model_performance}
\end{table}

\subsection{Plasticity Analysis}
To evaluate the model’s ability to adapt to new data, we record validation loss immediately after training on each task. Table~\ref{tab:plasticity_tab} reveals that larger models combined with replay and Reptile exhibit the strongest adaptation capabilities without sacrificing prior knowledge.

\begin{table}[h]
    \centering
    \renewcommand{\arraystretch}{1.2}
    \begin{tabular}{lccc|c}
        \toprule
        \textbf{Training Method} & $\mathcal{D}_0$ DCLM & $\mathcal{D}_1$ French & $\mathcal{D}_2$ German & \textbf{AVG} \\
        \midrule
        99M (No Replay) & 3.60 & 2.20 & 2.60 & 2.80 \\
        99M (25\% Replay) & 3.20 & 2.10 & 2.27 & 2.52 \\
        99M (50\% Replay) & 3.00 & 1.95 & 2.36 & 2.44 \\
        99M (Reptile Only) & 3.30 & 2.20 & 2.40 & 2.63 \\
        99M (25\% Replay + Reptile) & 3.10 & 2.00 & 2.29 & 2.46 \\
        99M (50\% Replay + Reptile) & 2.80 & 1.80 & 2.01 & 2.20 \\
        \hline
        560M (No Replay) & 3.30 & 1.90 & 2.40 & 2.53 \\
        560M (25\% Replay) & 2.90 & 1.65 & 1.91 & 2.15 \\
        560M (50\% Replay) & 2.55 & 1.30 & 2.34 & 2.06 \\
        560M (Reptile Only) & 3.10 & 1.85 & 2.30 & 2.42 \\
        560M (25\% Replay + Reptile) & 2.70 & 1.50 & 1.98 & 2.06 \\
        560M (50\% Replay + Reptile) & 2.35 & 1.10 & 1.60 & 1.68 \\
        \hline
        1B (No Replay) & 3.00 & 1.60 & 2.00 & 2.20 \\
        1B (25\% Replay) & 2.50 & 1.70 & 2.22 & 2.14 \\
        1B (50\% Replay) & 2.30 & 1.35 & 2.33 & 2.00 \\
        1B (Reptile Only) & 2.80 & 1.78 & 2.12 & 2.23 \\
        1B (25\% Replay + Reptile) & 2.55 & 1.45 & 2.28 & 2.09 \\
        1B (50\% Replay + Reptile) & 2.20 & 1.35 & 1.83 & 1.79 \\
        \hline
        6B (No Replay) & 1.90 & 1.10 & 1.10 & 1.37 \\
        6B (25\% Replay) & 1.20 & 0.90 & 0.90 & 1.00 \\
        6B (50\% Replay) & 1.10 & 0.70 & 0.80 & 0.87 \\
        6B (Reptile Only) & 1.50 & 1.00 & 1.09 & 1.20 \\
        6B (25\% Replay + Reptile) & 0.95 & 0.68 & 0.77 & 0.80 \\
        6B (50\% Replay + Reptile) & 0.90 & 0.65 & 0.74 & 0.76 \\
        \bottomrule
    \end{tabular}
    \caption{Learned validation loss comparison for individual tasks during CPT right after the model is trained on that task for the DCLM → French → German sequence. AVG represents the mean performance across the three datasets.}
    \label{tab:plasticity_tab}
\end{table}
\subsection{Scaling Laws and Tradeoffs}
\begin{figure}[h!]
    \centering
    \includegraphics[width=0.8\linewidth]{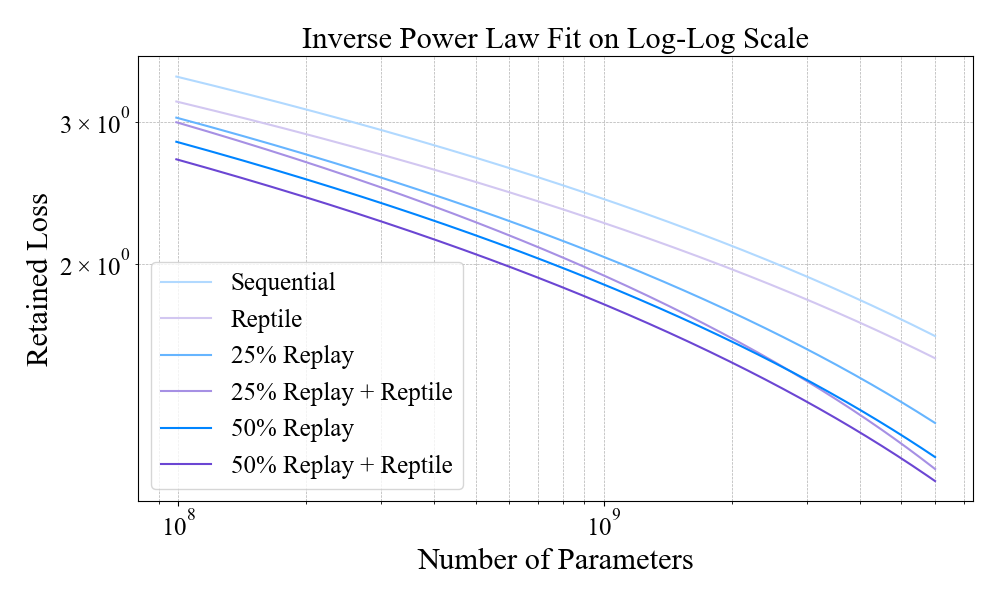}
    \caption{\textbf{Stability Scaling Analysis: Retained Loss vs. Model Size}. We consider the retained loss averaged across tasks at the end of training for each model and plot the results of an inverse power law fit for each model family. }
    \label{fig:scaling_plot_model_size}
\end{figure}

\begin{figure}[h!]
    \centering
    \includegraphics[width=0.8\linewidth]{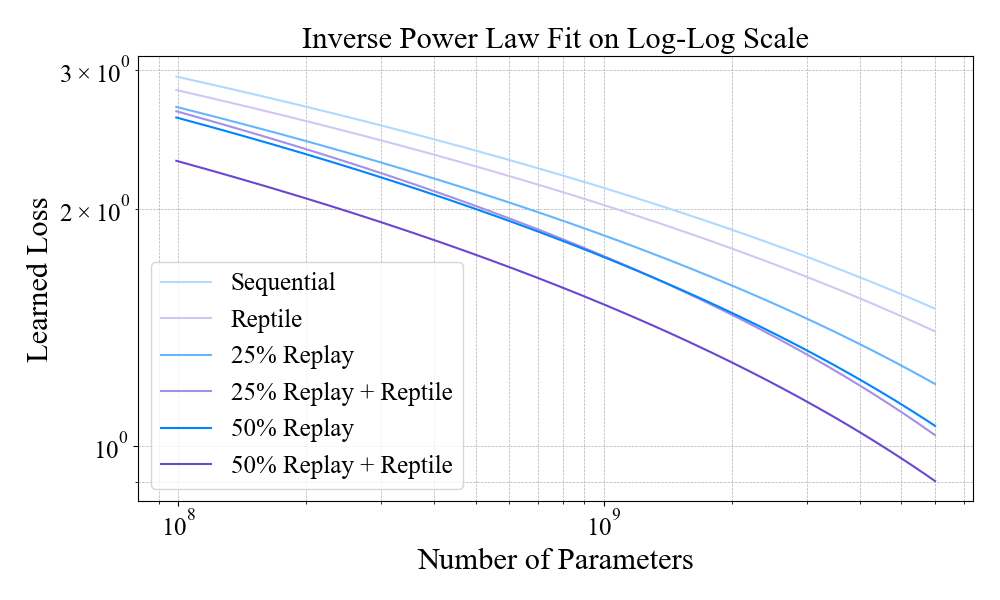}
    \caption{\textbf{Plasticity Scaling Analysis: Retained Loss vs. Model Size}. We consider the learned loss averaged across tasks at the end of training for each model and plot the results of an inverse power law fit for each model family. }
    \label{fig:scaling_plot_model_size_learned}
\end{figure}

Figures~\ref{fig:scaling_plot_model_size} and \ref{fig:scaling_plot_model_size_learned} visualize the scaling behavior of retained and learned loss versus model size. 

\clearpage
\section*{Additional Notes on Evaluation Metrics}

\paragraph{Note on Forgetting Metric for the First Task:} 
We acknowledge that the forgetting metric does not provide meaningful information for the first task (\(\mathcal{D}_0\)) in the continual pre-training sequence. We would like to clarify that the forgetting measure only becomes relevant starting from the second task onward, where it captures degradation in performance on earlier tasks due to the introduction of new data. 

Furthermore, we emphasize that traditional continual learning metrics such as forgetting are not directly applicable in our setting, as our pre-training is unsupervised and does not involve labeled data. As such, we define forgetting as the increase in validation loss relative to the best performance observed for each task. This definition, along with a formal description, has been added to the Evaluation Metrics section in the paper.

\clearpage
\section{Algorithms}
\section*{Algorithm 1: Experience Replay}
The basic experience replay mechanism interleaves a fixed proportion of replayed samples with fresh incoming data during training. This mechanism is described in Algorithm~\ref{ER}. \\
\textbf{Parameters:}
\begin{itemize}\setlength\itemsep{0pt}   
    \item $p$: Data‑generating distribution at step~$t$; each sample is drawn as $x_t \sim p(x \mid t)$.
    \item $\theta_0$: Initial network parameters before any continual‑learning updates.
    \item $N$: Total batch size per gradient step.
    \item $\alpha$: \emph{Replay ratio}; $\alpha N$ elements of every batch come from the memory buffer and $(1-\alpha)N$ are fresh samples.
    \item $T$: Number of training iterations (\texttt{for $t = 0,\,\dots,\,T{-}1$}).
    \item $M_t$: Replay memory after step~$t$ (initially $M_0 = \{\}$, then updated via reservoir sampling $M_{t+1} \leftarrow M_t \cup \{x\}$).
    \item $x_t$: Incoming data point received at training step~$t$.
    \item $B_M$: Mini‑batch of size $\alpha N$ sampled uniformly from $M_t$ (replay samples).
    \item $B$: Full training batch (size $N$) obtained by concatenating the newest $(1-\alpha)N$ examples with $B_M$.
    \item $\theta_t,\;\theta_{t+1}$: Model parameters immediately before and after the AdamW update computed on batch $B$.
\end{itemize}
\begin{algorithm}[H]
\caption{Experience Replay} \label{ER}
\begin{algorithmic}
  \State \Procedure{Train}{$p,\theta_0,N,\alpha$} 
  \State $M_0 \gets \{\}$
  \For{\texttt{$t = 0,...,T-1$}}
  \State // Receive incoming data point: 
  \State $x_t \sim p(x|t)$
  \State // Update every $(1-\alpha)N$ steps: 
  \If{$t \mod (1-\alpha)N = 0$}
  \State // Draw batch of size $\alpha N$ from buffer: 
  \State $B_M \gets sample(M_t,\alpha,N)$
  \State // Combine for full batch of size $N$:
  \State $B \gets x_t \cup ... \cup x_{t-(1-\alpha)N}  \cup B_M$
  \State // Update parameters with AdamW:
  \State $\theta_{t+1} \gets AdamW(B,\theta_t,\alpha)$
  \State // Reservoir sampling memory update:   
  \State $M_{t+1} \gets M_t  \cup \{x\}$
  \EndIf
  \EndFor
  \State \textbf{return} $\theta_{T},M_{T}$
  \EndProcedure
\end{algorithmic}
\end{algorithm}
\section*{Algorithm 2: Meta-Experience Replay}
We augment the basic replay procedure with Reptile-style meta-updates, promoting gradient alignment. The complete algorithm is shown below.
\\
\textbf{Parameters:}
\begin{itemize}\setlength\itemsep{0pt}
    \item $p$: Data‑generating distribution at step $t$; each incoming sample is drawn as $x_t \sim p(x\mid t)$.
    \item $\theta_0$: Initial model parameters before any continual‑training updates.
    \item $N$: Total batch size per gradient step (the combined batch $B$ always has $N$ examples).
    \item $\alpha$: \emph{Replay ratio}; $\alpha N$ items of every batch come from memory, leaving $(1-\alpha)N$ fresh examples.
    \item $k$: Number of batches between successive Reptile meta‑updates.
    \item $\epsilon$: Interpolation coefficient (meta‑learning rate) used in the Reptile update $\theta_t \leftarrow \theta_{t-k} + \epsilon(\theta_t-\theta_{t-k})$.
    \item $T$: Total number of training iterations 
    \item $M_t$: Replay memory after step $t$; updated via reservoir sampling $M_{t+1}\leftarrow M_t\cup\{x\}$.
    \item $x_t$: Incoming data point at step $t$. 
    \item $B_M$: Mini‑batch of size $\alpha N$ sampled uniformly from $M_t$ for replay. 
    \item $B$: Full training batch (size $N$) formed by concatenating the newest $(1-\alpha)N$ samples with $B_M$.
    \item $\theta_t,\;\theta_{t+1}$: Model parameters immediately before and after the AdamW update on batch $B$ (and $\theta_t$ is periodically interpolated via Reptile).
\end{itemize}

\begin{algorithm}[H]
\caption{Meta-Experience Replay} \label{MER}
\begin{algorithmic}
  \State \Procedure{Train}{$p,\theta_0,N,\alpha,k,\epsilon$} 
  \State $M_0 \gets \{\}$
  \For{\texttt{$t = 0,...,T-1$}}
  \State // Receive incoming data point: 
  \State $x_t \sim p(x|t)$
  \State // Update every $(1-\alpha)N$ steps: 
  \If{$t \mod (1-\alpha)N = 0$}
  \State // Draw batch of size $\alpha N$ from buffer: 
  \State $B_M \gets sample(M_t,\alpha,N)$
  \State // Combine for full batch of size $N$:
  \State $B \gets x_t \cup ... \cup x_{t-(1-\alpha)N}  \cup B_M$
  \State // Update parameters with AdamW:
  \State $\theta_{t+1} \gets AdamW(B,\theta_t,\alpha)$
  \State // Reservoir sampling memory update:   
  \State $M_{t+1} \gets M_t  \cup \{x\}$
  \State // Perform Reptile meta-update after every $k$ batches: 
  \If{$t \mod (1-\alpha)Nk = 0$}
  \State $\theta_{t} \gets \theta_{t-(1-\alpha)Nk} + \epsilon (\theta_{t} - \theta_{t-(1-\alpha)Nk})$
  \EndIf
  \EndIf
  \EndFor
  \State \textbf{return} $\theta_{T},M_{T}$
  \EndProcedure
\end{algorithmic}
\end{algorithm}

\section*{Algorithm 3: Disk-Based Replay Buffer with Prefetching}
To handle large-scale datasets efficiently, we implemented a disk-backed replay buffer with prefetching and queue-based sampling. This enables continual learning with effectively infinite memory capacity.
\\

\textbf{Parameters:}
\begin{itemize}
    \item $\mathcal{B}$: Buffer consisting of $F$ disk-backed files.
    \item $T$: Total buffer capacity in tokens.
    \item $S$: Size of each file in tokens.
    \item $L$: Sequence length (tokens per sample).
    \item $D$: Data directory path.
    \item $P$: Prefetch queue capacity.
    \item $X \in \mathbb{Z}^{N \times L}$: Input tensor batch of $N$ sequences.
    \item $B$: Batch size requested from buffer.
    \item $\beta$: Proportion of effective batch size to sample from buffer.
\end{itemize}

\begin{algorithm}[H]
\caption{ReplayBuffer$(T, S, L, D, P)$}
\label{alg:replay_buffer}
\begin{algorithmic}[1]

\Procedure{Init}{$T, S, L, D, P$}
    \State $F \gets \lfloor T / S \rfloor$  \Comment{Number of disk files}
    \State $\mathcal{B} \gets \texttt{[empty files]} \in D$, $\texttt{sizes} \gets \vec{0} \in \mathbb{N}^F$
    \If{\texttt{metadata.json} exists}
        \State \Call{LoadMetadata}{}
    \Else
        \State \Call{CreateFiles}{}
        \State \Call{Add}{$\mathbf{0} \in \mathbb{Z}^{1 \times L}$}  \Comment{Initialize metadata}
    \EndIf
    \State Launch background thread \Call{PrefetchFiles}{}
\EndProcedure

\Statex

\Function{Add}{$X \in \mathbb{Z}^{N \times L}$}
    \State Find $i$ where $\texttt{sizes}[i] + N \leq S$; else sample random $i$
    \State Compute offset $o \gets \texttt{sizes}[i] \cdot L \cdot s$ (with $s$: dtype size)
    \State Write $X$ to $\mathcal{B}[i]$ at offset $o$
    \State Update $\texttt{sizes}[i] \gets \texttt{sizes}[i] + N$, $\texttt{total} \gets \texttt{total} + N$
\EndFunction

\Statex

\Function{PrefetchFiles}{}
    \While{True}
        \If{$|\mathcal{Q}| < P$}
            \State Sample $i$ such that $\texttt{sizes}[i] > 0$
            \State Read raw bytes from $\mathcal{B}[i]$, reshape to $X \in \mathbb{Z}^{\texttt{sizes}[i] \times L}$
            \State $\mathcal{Q} \gets \mathcal{Q} \cup \{(X, i)\}$
        \Else
            \State \texttt{sleep}(0.1)
        \EndIf
    \EndWhile
\EndFunction

\Statex

\Function{GetBatch}{$\beta \in (0,1]$}
    \State $B \gets \lfloor \beta \cdot \texttt{eff\_batch\_size} \rfloor$
    \If{$\mathcal{Q} = \emptyset$} \Return $\emptyset$
    \Else
        \State Pop $(X, i)$ from $\mathcal{Q}$, sample $I \subset [1,\dots,|X|], |I| = B$
        \State \Return $X[I]$ on device
    \EndIf
\EndFunction

\Statex

\Function{SaveMetadata}{}
    \State Serialize $(T, S, L, \texttt{sizes}, \texttt{total}, \texttt{dtype}) \to$ JSON
\EndFunction

\Function{LoadMetadata}{}
    \State Load from JSON: restore buffer state and tensor properties
\EndFunction

\Function{CreateFiles}{}
    \For{$i = 1$ to $F$}
        \If{$\mathcal{B}[i]$ does not exist} create empty file \EndIf
    \EndFor
\EndFunction

\end{algorithmic}
\end{algorithm}

\end{document}

%% file: math_commands.tex
\usepackage{amsmath,amsfonts,bm}

\def\eqref#1{equation~\ref{#1}}

\def\1{\bm{1}}

\DeclareMathAlphabet{\mathsfit}{\encodingdefault}{\sfdefault}{m}{sl}
\SetMathAlphabet{\mathsfit}{bold}{\encodingdefault}{\sfdefault}{bx}{n}

\DeclareMathOperator*{\argmin}{arg\,min}

%% file: collas2025_conference.bbl
\begin{thebibliography}{67}
\providecommand{\natexlab}[1]{#1}
\providecommand{\url}[1]{\texttt{#1}}
\expandafter\ifx\csname urlstyle\endcsname\relax
  \providecommand{\doi}[1]{doi: #1}\else
  \providecommand{\doi}{doi: \begingroup \urlstyle{rm}\Url}\fi

\bibitem[Abadji et~al.(2022)Abadji, Suarez, Romary, and Sagot]{abadji2022cleanerdocumentorientedmultilingualcrawled}
Julien Abadji, Pedro~Ortiz Suarez, Laurent Romary, and Benoît Sagot.
\newblock Towards a cleaner document-oriented multilingual crawled corpus, 2022.
\newblock URL \url{https://arxiv.org/abs/2201.06642}.

\bibitem[Aloui et~al.(2024)Aloui, Chouikhi, Chaabane, Kchaou, and Dhaouadi]{aloui2024101billionarabicwords}
Manel Aloui, Hasna Chouikhi, Ghaith Chaabane, Haithem Kchaou, and Chehir Dhaouadi.
\newblock 101 billion arabic words dataset, 2024.
\newblock URL \url{https://arxiv.org/abs/2405.01590}.

\bibitem[Andonian et~al.(2023)Andonian, Anthony, Biderman, Black, Gali, Gao, Hallahan, Levy-Kramer, Leahy, Nestler, Parker, Pieler, Phang, Purohit, Schoelkopf, Stander, Songz, Tigges, Thérien, Wang, and Weinbach]{gpt-neox-library}
Alex Andonian, Quentin Anthony, Stella Biderman, Sid Black, Preetham Gali, Leo Gao, Eric Hallahan, Josh Levy-Kramer, Connor Leahy, Lucas Nestler, Kip Parker, Michael Pieler, Jason Phang, Shivanshu Purohit, Hailey Schoelkopf, Dashiell Stander, Tri Songz, Curt Tigges, Benjamin Thérien, Phil Wang, and Samuel Weinbach.
\newblock {GPT-NeoX: Large Scale Autoregressive Language Modeling in PyTorch}, 9 2023.
\newblock URL \url{https://www.github.com/eleutherai/gpt-neox}.

\bibitem[Bashivan et~al.(2019)Bashivan, Schrimpf, Ajemian, Rish, Riemer, and Tu]{bashivan2019continual}
Pouya Bashivan, Martin Schrimpf, Robert Ajemian, Irina Rish, Matthew Riemer, and Yuhai Tu.
\newblock Continual learning with self-organizing maps.
\newblock \emph{arXiv preprint arXiv:1904.09330}, 2019.

\bibitem[Bisk et~al.(2019)Bisk, Zellers, Bras, Gao, and Choi]{bisk2019piqareasoningphysicalcommonsense}
Yonatan Bisk, Rowan Zellers, Ronan~Le Bras, Jianfeng Gao, and Yejin Choi.
\newblock Piqa: Reasoning about physical commonsense in natural language, 2019.
\newblock URL \url{https://arxiv.org/abs/1911.11641}.

\bibitem[Carpenter \& Grossberg(1987)Carpenter and Grossberg]{StabilityPlasticity}
Gail~A Carpenter and Stephen Grossberg.
\newblock A massively parallel architecture for a self-organizing neural pattern recognition machine.
\newblock \emph{Computer vision, graphics, and image processing}, 37\penalty0 (1):\penalty0 54--115, 1987.

\bibitem[Cases et~al.(2019)Cases, Rosenbaum, Riemer, Geiger, Klinger, Tamkin, Li, Agarwal, Greene, Jurafsky, et~al.]{cases2019recursive}
Ignacio Cases, Clemens Rosenbaum, Matthew Riemer, Atticus Geiger, Tim Klinger, Alex Tamkin, Olivia Li, Sandhini Agarwal, Joshua~D Greene, Dan Jurafsky, et~al.
\newblock Recursive routing networks: Learning to compose modules for language understanding.
\newblock In \emph{Proceedings of the 2019 Conference of the North American Chapter of the Association for Computational Linguistics: Human Language Technologies, Volume 1 (Long and Short Papers)}, pp.\  3631--3648, 2019.

\bibitem[Chang et~al.(2019)Chang, Gupta, Levine, and Griffiths]{chang2018automatically}
Michael~B Chang, Abhishek Gupta, Sergey Levine, and Thomas~L Griffiths.
\newblock Automatically composing representation transformations as a means for generalization.
\newblock 2019.

\bibitem[Chaudhry et~al.(2019{\natexlab{a}})Chaudhry, Ranzato, Rohrbach, and Elhoseiny]{agem}
Arslan Chaudhry, Marc'Aurelio Ranzato, Marcus Rohrbach, and Mohamed Elhoseiny.
\newblock Efficient lifelong learning with a-gem.
\newblock \emph{International Conference on Learning Representations}, 2019{\natexlab{a}}.

\bibitem[Chaudhry et~al.(2019{\natexlab{b}})Chaudhry, Rohrbach, Elhoseiny, Ajanthan, Dokania, Torr, and Ranzato]{chaudhry2019tinyepisodicmemoriescontinual}
Arslan Chaudhry, Marcus Rohrbach, Mohamed Elhoseiny, Thalaiyasingam Ajanthan, Puneet~K. Dokania, Philip H.~S. Torr, and Marc'Aurelio Ranzato.
\newblock On tiny episodic memories in continual learning, 2019{\natexlab{b}}.
\newblock URL \url{https://arxiv.org/abs/1902.10486}.

\bibitem[Chen et~al.(2024)Chen, Chen, Wang, Zhou, Zhu, Jiang, Min, Zhao, Dou, Mao, Lin, Song, Xu, Chen, Yan, Wei, Hu, Huang, and Wen]{chen2024effectiveefficientcontinualpretraining}
Jie Chen, Zhipeng Chen, Jiapeng Wang, Kun Zhou, Yutao Zhu, Jinhao Jiang, Yingqian Min, Wayne~Xin Zhao, Zhicheng Dou, Jiaxin Mao, Yankai Lin, Ruihua Song, Jun Xu, Xu~Chen, Rui Yan, Zhewei Wei, Di~Hu, Wenbing Huang, and Ji-Rong Wen.
\newblock Towards effective and efficient continual pre-training of large language models, 2024.
\newblock URL \url{https://arxiv.org/abs/2407.18743}.

\bibitem[Chen et~al.(2023)Chen, Zhou, Du, Huang, Laudon, Chen, and Cui]{chen2023lifelong}
Wuyang Chen, Yanqi Zhou, Nan Du, Yanping Huang, James Laudon, Zhifeng Chen, and Claire Cui.
\newblock Lifelong language pretraining with distribution-specialized experts.
\newblock In \emph{International Conference on Machine Learning}, pp.\  5383--5395. PMLR, 2023.

\bibitem[Davis \& Arel(2013)Davis and Arel]{davis2013low}
Andrew Davis and Itamar Arel.
\newblock Low-rank approximations for conditional feedforward computation in deep neural networks.
\newblock \emph{arXiv preprint arXiv:1312.4461}, 2013.

\bibitem[De~Lange et~al.(2021)De~Lange, Aljundi, Masana, Parisot, Jia, Leonardis, Slabaugh, and Tuytelaars]{de2021continual}
Matthias De~Lange, Rahaf Aljundi, Marc Masana, Sarah Parisot, Xu~Jia, Ale{\v{s}} Leonardis, Gregory Slabaugh, and Tinne Tuytelaars.
\newblock A continual learning survey: Defying forgetting in classification tasks.
\newblock \emph{IEEE transactions on pattern analysis and machine intelligence}, 44\penalty0 (7):\penalty0 3366--3385, 2021.

\bibitem[Duwal et~al.(2024)Duwal, Prasai, and Manandhar]{duwal2024domainadaptativecontinuallearninglowresource}
Sharad Duwal, Suraj Prasai, and Suresh Manandhar.
\newblock Domain-adaptative continual learning for low-resource tasks: Evaluation on nepali, 2024.
\newblock URL \url{https://arxiv.org/abs/2412.13860}.

\bibitem[Fernandez et~al.(2024)Fernandez, Bisk, and Strubell]{fernandez2024gradientlocalizationimproveslifelong}
Jared Fernandez, Yonatan Bisk, and Emma Strubell.
\newblock Gradient localization improves lifelong pretraining of language models, 2024.
\newblock URL \url{https://arxiv.org/abs/2411.04448}.

\bibitem[Fernando et~al.(2017)Fernando, Banarse, Blundell, Zwols, Ha, Rusu, Pritzel, and Wierstra]{Pathnet}
Chrisantha Fernando, Dylan Banarse, Charles Blundell, Yori Zwols, David Ha, Andrei~A Rusu, Alexander Pritzel, and Daan Wierstra.
\newblock Pathnet: Evolution channels gradient descent in super neural networks.
\newblock \emph{arXiv preprint arXiv:1701.08734}, 2017.

\bibitem[Finn et~al.(2017)Finn, Abbeel, and Levine]{maml}
Chelsea Finn, Pieter Abbeel, and Sergey Levine.
\newblock Model-agnostic meta-learning for fast adaptation of deep networks.
\newblock In \emph{International conference on machine learning}, pp.\  1126--1135. PMLR, 2017.

\bibitem[Fujii et~al.(2024)Fujii, Nakamura, Loem, Iida, Ohi, Hattori, Shota, Mizuki, Yokota, and Okazaki]{fujii2024continual}
Kazuki Fujii, Taishi Nakamura, Mengsay Loem, Hiroki Iida, Masanari Ohi, Kakeru Hattori, Hirai Shota, Sakae Mizuki, Rio Yokota, and Naoaki Okazaki.
\newblock Continual pre-training for cross-lingual llm adaptation: Enhancing japanese language capabilities.
\newblock \emph{arXiv preprint arXiv:2404.17790}, 2024.

\bibitem[Gogoulou et~al.(2024)Gogoulou, Lesort, Boman, and Nivre]{gogoulou2024continuallearninglanguageshift}
Evangelia Gogoulou, Timothée Lesort, Magnus Boman, and Joakim Nivre.
\newblock Continual learning under language shift, 2024.
\newblock URL \url{https://arxiv.org/abs/2311.01200}.

\bibitem[Gupta et~al.(2023)Gupta, Thérien, Ibrahim, Richter, Anthony, Belilovsky, Rish, and Lesort]{gupta2023continualpretraininglargelanguage}
Kshitij Gupta, Benjamin Thérien, Adam Ibrahim, Mats~L. Richter, Quentin Anthony, Eugene Belilovsky, Irina Rish, and Timothée Lesort.
\newblock Continual pre-training of large language models: How to (re)warm your model?, 2023.
\newblock URL \url{https://arxiv.org/abs/2308.04014}.

\bibitem[Gururangan et~al.(2021)Gururangan, Lewis, Holtzman, Smith, and Zettlemoyer]{gururangan2021demix}
Suchin Gururangan, Mike Lewis, Ari Holtzman, Noah~A Smith, and Luke Zettlemoyer.
\newblock Demix layers: Disentangling domains for modular language modeling.
\newblock \emph{arXiv preprint arXiv:2108.05036}, 2021.

\bibitem[Hattori(2024)]{abeja_cc_ja_techblog2024}
Kyo Hattori.
\newblock Building a large-scale japanese corpus from common crawl and its preprocessing.
\newblock Online; ABEJA Tech Blog, May 2024.
\newblock \url{https://tech-blog.abeja.asia/entry/abeja-nedo-project-part2-202405}.

\bibitem[Ibrahim et~al.(2024)Ibrahim, Th{\'e}rien, Gupta, Richter, Anthony, Lesort, Belilovsky, and Rish]{ibrahim2024simple}
Adam Ibrahim, Benjamin Th{\'e}rien, Kshitij Gupta, Mats~L Richter, Quentin Anthony, Timoth{\'e}e Lesort, Eugene Belilovsky, and Irina Rish.
\newblock Simple and scalable strategies to continually pre-train large language models.
\newblock \emph{arXiv preprint arXiv:2403.08763}, 2024.

\bibitem[Jacobs et~al.(1991)Jacobs, Jordan, Nowlan, and Hinton]{Hinton91}
Robert~A Jacobs, Michael~I Jordan, Steven~J Nowlan, and Geoffrey~E Hinton.
\newblock Adaptive mixtures of local experts.
\newblock \emph{Neural computation}, 3\penalty0 (1):\penalty0 79--87, 1991.

\bibitem[Jin et~al.(2019)Jin, Dhingra, Liu, Cohen, and Lu]{jin2019pubmedqadatasetbiomedicalresearch}
Qiao Jin, Bhuwan Dhingra, Zhengping Liu, William~W. Cohen, and Xinghua Lu.
\newblock Pubmedqa: A dataset for biomedical research question answering, 2019.
\newblock URL \url{https://arxiv.org/abs/1909.06146}.

\bibitem[Jin et~al.(2021)Jin, Zhang, Zhu, Xiao, Li, Wei, Arnold, and Ren]{jin2021lifelong}
Xisen Jin, Dejiao Zhang, Henghui Zhu, Wei Xiao, Shang-Wen Li, Xiaokai Wei, Andrew Arnold, and Xiang Ren.
\newblock Lifelong pretraining: Continually adapting language models to emerging corpora.
\newblock \emph{arXiv preprint arXiv:2110.08534}, 2021.

\bibitem[Jordan \& Jacobs(1994)Jordan and Jacobs]{Jordan94}
Michael~I Jordan and Robert~A Jacobs.
\newblock Hierarchical mixtures of experts and the em algorithm.
\newblock \emph{Neural computation}, 6\penalty0 (2):\penalty0 181--214, 1994.

\bibitem[Kaushal et~al.(2024)Kaushal, Vaidhya, Mondal, Pandey, Bhagat, and Rish]{kaushal2024spectrasurprisingeffectivenesspretraining}
Ayush Kaushal, Tejas Vaidhya, Arnab~Kumar Mondal, Tejas Pandey, Aaryan Bhagat, and Irina Rish.
\newblock Spectra: Surprising effectiveness of pretraining ternary language models at scale, 2024.
\newblock URL \url{https://arxiv.org/abs/2407.12327}.

\bibitem[Ke et~al.(2023)Ke, Shao, Lin, Konishi, Kim, and Liu]{ke2023continual}
Zixuan Ke, Yijia Shao, Haowei Lin, Tatsuya Konishi, Gyuhak Kim, and Bing Liu.
\newblock Continual pre-training of language models.
\newblock \emph{arXiv preprint arXiv:2302.03241}, 2023.

\bibitem[Khetarpal et~al.(2022)Khetarpal, Riemer, Rish, and Precup]{khetarpal2022towards}
Khimya Khetarpal, Matthew Riemer, Irina Rish, and Doina Precup.
\newblock Towards continual reinforcement learning: A review and perspectives.
\newblock \emph{Journal of Artificial Intelligence Research}, 75:\penalty0 1401--1476, 2022.

\bibitem[Kostas et~al.(2020)Kostas, Nota, and Thomas]{kostas20a}
James Kostas, Chris Nota, and Philip Thomas.
\newblock Asynchronous coagent networks.
\newblock In Hal~Daumé III and Aarti Singh (eds.), \emph{Proceedings of the 37th International Conference on Machine Learning}, volume 119 of \emph{Proceedings of Machine Learning Research}, pp.\  5426--5435. PMLR, 13--18 Jul 2020.
\newblock URL \url{https://proceedings.mlr.press/v119/kostas20a.html}.

\bibitem[Li \& Lee(2024)Li and Lee]{li2024examining}
Chen-An Li and Hung-Yi Lee.
\newblock Examining forgetting in continual pre-training of aligned large language models.
\newblock \emph{arXiv preprint arXiv:2401.03129}, 2024.

\bibitem[Li et~al.(2024)Li, Fang, Smyrnis, Ivgi, Jordan, Gadre, Bansal, Guha, Keh, Arora, Garg, Xin, Muennighoff, Heckel, Mercat, Chen, Gururangan, Wortsman, Albalak, Bitton, Nezhurina, Abbas, Hsieh, Ghosh, Gardner, Kilian, Zhang, Shao, Pratt, Sanyal, Ilharco, Daras, Marathe, Gokaslan, Zhang, Chandu, Nguyen, Vasiljevic, Kakade, Song, Sanghavi, Faghri, Oh, Zettlemoyer, Lo, El-Nouby, Pouransari, Toshev, Wang, Groeneveld, Soldaini, Koh, Jitsev, Kollar, Dimakis, Carmon, Dave, Schmidt, and Shankar]{li2024datacomplmsearchgenerationtraining}
Jeffrey Li, Alex Fang, Georgios Smyrnis, Maor Ivgi, Matt Jordan, Samir Gadre, Hritik Bansal, Etash Guha, Sedrick Keh, Kushal Arora, Saurabh Garg, Rui Xin, Niklas Muennighoff, Reinhard Heckel, Jean Mercat, Mayee Chen, Suchin Gururangan, Mitchell Wortsman, Alon Albalak, Yonatan Bitton, Marianna Nezhurina, Amro Abbas, Cheng-Yu Hsieh, Dhruba Ghosh, Josh Gardner, Maciej Kilian, Hanlin Zhang, Rulin Shao, Sarah Pratt, Sunny Sanyal, Gabriel Ilharco, Giannis Daras, Kalyani Marathe, Aaron Gokaslan, Jieyu Zhang, Khyathi Chandu, Thao Nguyen, Igor Vasiljevic, Sham Kakade, Shuran Song, Sujay Sanghavi, Fartash Faghri, Sewoong Oh, Luke Zettlemoyer, Kyle Lo, Alaaeldin El-Nouby, Hadi Pouransari, Alexander Toshev, Stephanie Wang, Dirk Groeneveld, Luca Soldaini, Pang~Wei Koh, Jenia Jitsev, Thomas Kollar, Alexandros~G. Dimakis, Yair Carmon, Achal Dave, Ludwig Schmidt, and Vaishaal Shankar.
\newblock Datacomp-lm: In search of the next generation of training sets for language models, 2024.
\newblock URL \url{https://arxiv.org/abs/2406.11794}.

\bibitem[Lin(1992)]{Lin92}
Long-Ji Lin.
\newblock Self-improving reactive agents based on reinforcement learning, planning and teaching.
\newblock \emph{Machine learning}, 8\penalty0 (3-4):\penalty0 293--321, 1992.

\bibitem[Lopez{-}Paz \& Ranzato(2017)Lopez{-}Paz and Ranzato]{gem}
David Lopez{-}Paz and Marc'Aurelio Ranzato.
\newblock Gradient episodic memory for continuum learning.
\newblock \emph{NIPS}, 2017.

\bibitem[McCloskey \& Cohen(1989)McCloskey and Cohen]{CF}
Michael McCloskey and Neal~J Cohen.
\newblock Catastrophic interference in connectionist networks: The sequential learning problem.
\newblock \emph{Psychology of learning and motivation}, 24:\penalty0 109--165, 1989.

\bibitem[Miikkulainen(1993)]{Miikkulainen1993}
Risto Miikkulainen.
\newblock \emph{Subsymbolic natural language processing - an integrated model of scripts, lexicon, and memory}.
\newblock Neural network modeling and connectionism. {MIT} Press, 1993.
\newblock ISBN 978-0-262-13290-9.

\bibitem[Mnih et~al.(2015)Mnih, Kavukcuoglu, Silver, Rusu, Veness, Bellemare, Graves, Riedmiller, Fidjeland, Ostrovski, et~al.]{dqn}
Volodymyr Mnih, Koray Kavukcuoglu, David Silver, Andrei~A Rusu, Joel Veness, Marc~G Bellemare, Alex Graves, Martin Riedmiller, Andreas~K Fidjeland, Georg Ostrovski, et~al.
\newblock Human-level control through deep reinforcement learning.
\newblock \emph{Nature}, 518\penalty0 (7540):\penalty0 529--533, 2015.

\bibitem[Murre(1992)]{Murre92}
Jacob~MJ Murre.
\newblock \emph{Learning and categorization in modular neural networks.}
\newblock Lawrence Erlbaum Associates, Inc, 1992.

\bibitem[Nichol et~al.(2018)Nichol, Achiam, and Schulman]{nichol2018firstordermetalearningalgorithms}
Alex Nichol, Joshua Achiam, and John Schulman.
\newblock On first-order meta-learning algorithms, 2018.
\newblock URL \url{https://arxiv.org/abs/1803.02999}.

\bibitem[Rebuffi et~al.(2017)Rebuffi, Kolesnikov, Sperl, and Lampert]{icarl}
Sylvestre-Alvise Rebuffi, Alexander Kolesnikov, Georg Sperl, and Christoph~H Lampert.
\newblock icarl: Incremental classifier and representation learning.
\newblock In \emph{Proceedings of the IEEE conference on Computer Vision and Pattern Recognition}, pp.\  2001--2010, 2017.

\bibitem[Riemer et~al.(2017{\natexlab{a}})Riemer, Franceschini, Bouneffouf, and Klinger]{GenerativeDist}
Matthew Riemer, Michele Franceschini, Djallel Bouneffouf, and Tim Klinger.
\newblock Generative knowledge distillation for general purpose function compression.
\newblock \emph{NIPS 2017 Workshop on Teaching Machines, Robots, and Humans}, 5:\penalty0 30, 2017{\natexlab{a}}.

\bibitem[Riemer et~al.(2017{\natexlab{b}})Riemer, Klinger, Franceschini, and Bouneffouf]{Recollections}
Matthew Riemer, Tim Klinger, Michele Franceschini, and Djallel Bouneffouf.
\newblock Scalable recollections for continual lifelong learning.
\newblock \emph{arXiv preprint arXiv:1711.06761}, 2017{\natexlab{b}}.

\bibitem[Riemer et~al.(2019{\natexlab{a}})Riemer, Cases, Ajemian, Liu, Rish, Tu, and Tesauro]{riemer2019learninglearnforgettingmaximizing}
Matthew Riemer, Ignacio Cases, Robert Ajemian, Miao Liu, Irina Rish, Yuhai Tu, and Gerald Tesauro.
\newblock Learning to learn without forgetting by maximizing transfer and minimizing interference.
\newblock In \emph{International Conference on Learning Representations}, 2019{\natexlab{a}}.

\bibitem[Riemer et~al.(2019{\natexlab{b}})Riemer, Klinger, Bouneffouf, and Franceschini]{riemer2019scalable}
Matthew Riemer, Tim Klinger, Djallel Bouneffouf, and Michele Franceschini.
\newblock Scalable recollections for continual lifelong learning.
\newblock In \emph{Proceedings of the AAAI Conference on Artificial Intelligence}, volume~33, pp.\  1352--1359, 2019{\natexlab{b}}.

\bibitem[Riemer et~al.(2022)Riemer, Raparthy, Cases, Subbaraj, Touzel, and Rish]{polynomial}
Matthew Riemer, Sharath~Chandra Raparthy, Ignacio Cases, Gopeshh Subbaraj, Maximilian~Puelma Touzel, and Irina Rish.
\newblock Continual learning in environments with polynomial mixing times.
\newblock \emph{Advances in Neural Information Processing Systems}, 2022.

\bibitem[Riemer et~al.(2024)Riemer, Khetarpal, Rajendran, and Chandar]{riemerbalancing}
Matthew Riemer, Khimya Khetarpal, Janarthanan Rajendran, and Sarath Chandar.
\newblock Balancing context length and mixing times for reinforcement learning at scale.
\newblock In \emph{The Thirty-eighth Annual Conference on Neural Information Processing Systems}, 2024.

\bibitem[Ring(1994)]{Ring94}
Mark~Bishop Ring.
\newblock \emph{Continual learning in reinforcement environments}.
\newblock PhD thesis, University of Texas at Austin Austin, Texas 78712, 1994.

\bibitem[Robins(1995)]{Robins95}
Anthony Robins.
\newblock Catastrophic forgetting, rehearsal and pseudorehearsal.
\newblock \emph{Connection Science}, 7\penalty0 (2):\penalty0 123--146, 1995.

\bibitem[Rosenbaum et~al.(2018)Rosenbaum, Klinger, and Riemer]{rosenbaum2018routing}
Clemens Rosenbaum, Tim Klinger, and Matthew Riemer.
\newblock Routing networks: Adaptive selection of non-linear functions for multi-task learning.
\newblock In \emph{International Conference on Learning Representations}. International Conference on Learning Representations, ICLR, 2018.

\bibitem[Rosenbaum et~al.(2019{\natexlab{a}})Rosenbaum, Cases, Riemer, Geiger, Karttunen, Greene, Jurafsky, and Potts]{rosenbaum2019dispatched}
Clemens Rosenbaum, Ignacio Cases, Matthew Riemer, Atticus Geiger, Lauri Karttunen, Joshua~D Greene, Dan Jurafsky, and Christopher Potts.
\newblock Dispatched routing networks.
\newblock Technical report, 2019{\natexlab{a}}.

\bibitem[Rosenbaum et~al.(2019{\natexlab{b}})Rosenbaum, Cases, Riemer, and Klinger]{rosenbaum2019routing}
Clemens Rosenbaum, Ignacio Cases, Matthew Riemer, and Tim Klinger.
\newblock Routing networks and the challenges of modular and compositional computation.
\newblock \emph{arXiv preprint arXiv:1904.12774}, 2019{\natexlab{b}}.

\bibitem[Roth et~al.(2024)Roth, Udandarao, Dziadzio, Prabhu, Cherti, Vinyals, H{\'e}naff, Albanie, Bethge, and Akata]{roth2024practitioner}
Karsten Roth, Vishaal Udandarao, Sebastian Dziadzio, Ameya Prabhu, Mehdi Cherti, Oriol Vinyals, Olivier H{\'e}naff, Samuel Albanie, Matthias Bethge, and Zeynep Akata.
\newblock A practitioner's guide to continual multimodal pretraining.
\newblock \emph{arXiv preprint arXiv:2408.14471}, 2024.

\bibitem[Shazeer et~al.(2017)Shazeer, Mirhoseini, Maziarz, Davis, Le, Hinton, and Dean]{largeneuralnets}
Noam Shazeer, Azalia Mirhoseini, Krzysztof Maziarz, Andy Davis, Quoc~V. Le, Geoffrey~E. Hinton, and Jeff Dean.
\newblock Outrageously large neural networks: The sparsely-gated mixture-of-experts layer.
\newblock \emph{CoRR}, abs/1701.06538, 2017.
\newblock URL \url{http://arxiv.org/abs/1701.06538}.

\bibitem[Shi et~al.(2024)Shi, Xu, Wang, Qin, Wang, Wang, Wang, Ebrahimi, and Wang]{shi2024continual}
Haizhou Shi, Zihao Xu, Hengyi Wang, Weiyi Qin, Wenyuan Wang, Yibin Wang, Zifeng Wang, Sayna Ebrahimi, and Hao Wang.
\newblock Continual learning of large language models: A comprehensive survey.
\newblock \emph{ACM Computing Surveys}, 2024.

\bibitem[Su{\'a}rez et~al.(2019)Su{\'a}rez, Sagot, and Romary]{suarez2019asynchronous}
Pedro Javier~Ortiz Su{\'a}rez, Beno{\^\i}t Sagot, and Laurent Romary.
\newblock Asynchronous pipeline for processing huge corpora on medium to low resource infrastructures.
\newblock In \emph{7th Workshop on the Challenges in the Management of Large Corpora (CMLC-7)}. Leibniz-Institut f{\"u}r Deutsche Sprache, 2019.

\bibitem[Su{\'a}rez et~al.(2020)Su{\'a}rez, Romary, and Sagot]{suarez2020monolingual}
Pedro Javier~Ortiz Su{\'a}rez, Laurent Romary, and Beno{\^\i}t Sagot.
\newblock A monolingual approach to contextualized word embeddings for mid-resource languages.
\newblock \emph{arXiv preprint arXiv:2006.06202}, 2020.

\bibitem[Th{\'e}rien et~al.(2025)Th{\'e}rien, Joseph, Sarwar, Panda, Das, Zhang, Rawls, Sahu, Belilovsky, and Rish]{therien2025continual}
Benjamin Th{\'e}rien, Charles-{\'E}tienne Joseph, Zain Sarwar, Ashwinee Panda, Anirban Das, Shi-Xiong Zhang, Stephen Rawls, Sambit Sahu, Eugene Belilovsky, and Irina Rish.
\newblock Continual pre-training of moes: How robust is your router?
\newblock \emph{arXiv preprint arXiv:2503.05029}, 2025.

\bibitem[Thomas(2011)]{thomas}
Philip~S. Thomas.
\newblock Policy gradient coagent networks.
\newblock In J.~Shawe-Taylor, R.~Zemel, P.~Bartlett, F.~Pereira, and K.Q. Weinberger (eds.), \emph{Advances in Neural Information Processing Systems}, volume~24. Curran Associates, Inc., 2011.
\newblock URL \url{https://proceedings.neurips.cc/paper_files/paper/2011/file/1e6e0a04d20f50967c64dac2d639a577-Paper.pdf}.

\bibitem[Thrun(1994)]{Thrun94}
Sebastian Thrun.
\newblock Lifelong learning perspective for mobile robot control.
\newblock In \emph{Proceedings of the IEEE/RSJ/GI International Conference on Intelligent Robots and Systems}, volume~1, pp.\  23--30, 1994.

\bibitem[Vitter(1985)]{RS}
Jeffrey~S Vitter.
\newblock Random sampling with a reservoir.
\newblock \emph{ACM Transactions on Mathematical Software (TOMS)}, 11\penalty0 (1):\penalty0 37--57, 1985.

\bibitem[Yu et~al.(2020)Yu, Kumar, Gupta, Levine, Hausman, and Finn]{pcgrad}
Tianhe Yu, Saurabh Kumar, Abhishek Gupta, Sergey Levine, Karol Hausman, and Chelsea Finn.
\newblock Gradient surgery for multi-task learning.
\newblock \emph{Advances in neural information processing systems}, 33:\penalty0 5824--5836, 2020.

\bibitem[Zellers et~al.(2019)Zellers, Holtzman, Bisk, Farhadi, and Choi]{zellers2019hellaswagmachinereallyfinish}
Rowan Zellers, Ari Holtzman, Yonatan Bisk, Ali Farhadi, and Yejin Choi.
\newblock Hellaswag: Can a machine really finish your sentence?, 2019.
\newblock URL \url{https://arxiv.org/abs/1905.07830}.

\bibitem[Zhang et~al.(2019)Zhang, Lucas, Ba, and Hinton]{zhang2019lookahead}
Michael Zhang, James Lucas, Jimmy Ba, and Geoffrey~E Hinton.
\newblock Lookahead optimizer: k steps forward, 1 step back.
\newblock \emph{Advances in neural information processing systems}, 32, 2019.

\bibitem[Zhou et~al.(2024)Zhou, Sun, Ning, Ye, and Zhan]{zhou2024continual}
Da-Wei Zhou, Hai-Long Sun, Jingyi Ning, Han-Jia Ye, and De-Chuan Zhan.
\newblock Continual learning with pre-trained models: A survey.
\newblock \emph{arXiv preprint arXiv:2401.16386}, 2024.

\bibitem[Zini et~al.(2020)Zini, Pedramfar, Riemer, Moradipari, and Liu]{zini2020coagent}
Modjtaba~Shokrian Zini, Mohammad Pedramfar, Matthew Riemer, Ahmadreza Moradipari, and Miao Liu.
\newblock Coagent networks revisited.
\newblock \emph{arXiv preprint arXiv:2001.10474}, 2020.

\end{thebibliography}
